\documentclass[10pt,twocolumn,letterpaper]{article}

\usepackage[table]{xcolor}

\usepackage[pagenumbers]{cvpr} 

\usepackage{multirow} 

\definecolor{cvprblue}{rgb}{0.21,0.49,0.74}
\definecolor{grey}{rgb}{0.5,0.5,0.5}
\usepackage[pagebackref,breaklinks,colorlinks,allcolors=cvprblue]{hyperref}
\usepackage{mmstyle}

\title{TokenHSI: Unified Synthesis of Physical Human-Scene Interactions \\ through Task Tokenization}

\author{Liang Pan$^{1,2}$ \quad  Zeshi Yang$^{3}$ \quad  Zhiyang Dou$^2$ \quad  Wenjia Wang$^{2}$ \quad  Buzhen Huang$^4$ \\[0.8mm] Bo Dai$^{2,5}$ \quad  Taku Komura$^{2}$ \quad  Jingbo Wang$^{1\dagger}$ \\ 
\normalsize $^1$ Shanghai AI Laboratory \quad
\normalsize $^2$ The University of Hong Kong \quad
\normalsize $^3$ Independent Researcher \quad \\
\normalsize $^4$ Southeast University \quad
\normalsize $^5$ Feeling AI \quad \\
\normalsize \href{https://liangpan99.github.io/TokenHSI}{\textbf{https://liangpan99.github.io/TokenHSI}}
}

\begin{document}

\newcommand{\ours}{TokenHSI\xspace}

\twocolumn[{
\renewcommand\twocolumn[1][]{#1}
\maketitle

\begin{center}
    \centering
    \captionsetup{type=figure}
    \includegraphics[width=1.0\textwidth]{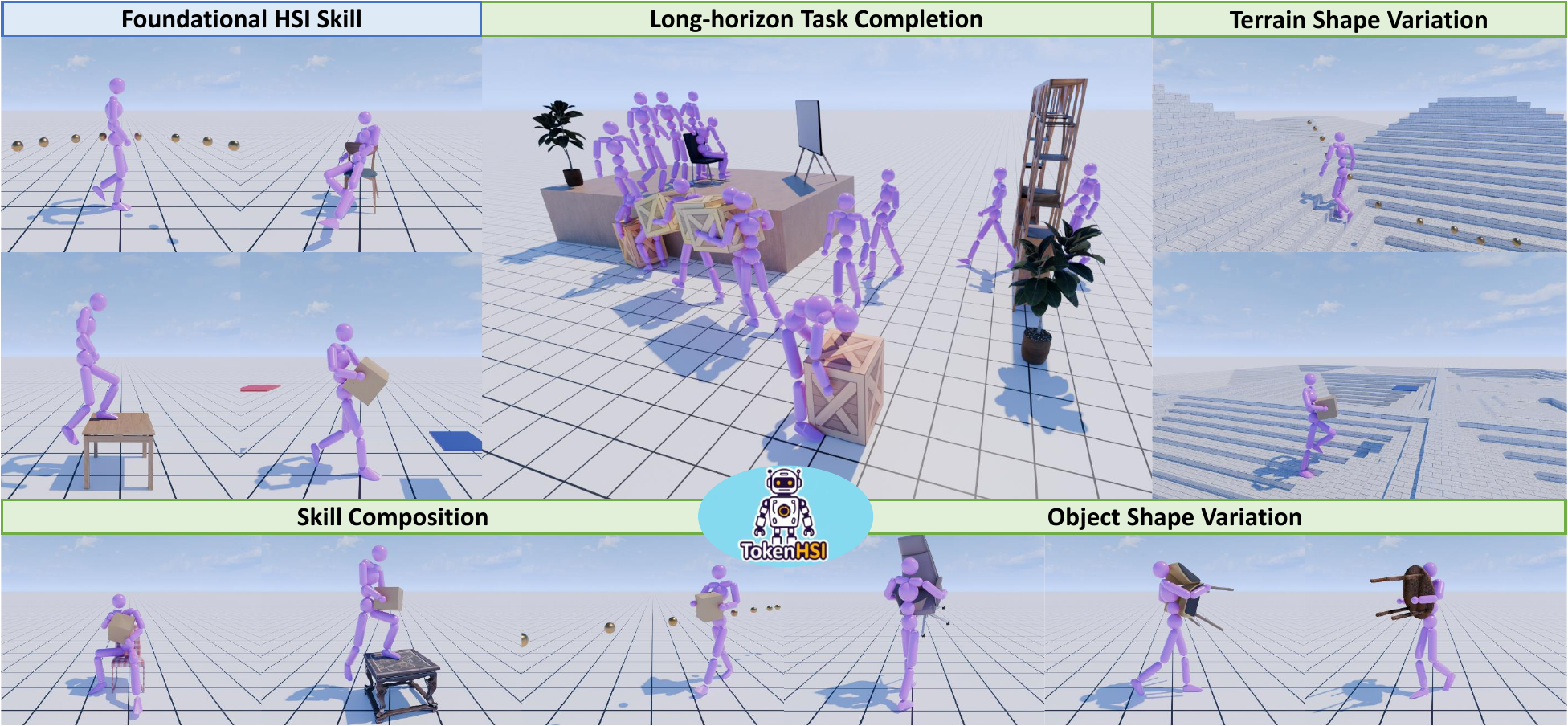}
    \captionof{figure}{Introducing \textbf{\ours}, a unified model that enables physics-based characters to perform diverse human-scene interaction tasks. It excels at seamlessly unifying multiple \textbf{\textcolor[RGB]{68, 114, 196}{foundational HSI skills}} within a single transformer network and flexibly adapting learned skills to \textbf{\textcolor[RGB]{112, 180, 71}{challenging new tasks}}, including skill composition, object/terrain shape variation, and long-horizon task completion.}
    \label{fig:teaser}
\end{center}
}]

\renewcommand{\thefootnote}{\fnsymbol{footnote}}
\footnotetext[2]{Corresponding author.}

\begin{abstract}
Synthesizing diverse and physically plausible Human-Scene Interactions~(HSI) is pivotal for both computer animation and embodied AI. Despite encouraging progress, current methods mainly focus on developing separate controllers, each specialized for a specific interaction task. This significantly hinders the ability to tackle a wide variety of challenging HSI tasks that require the integration of multiple skills, \eg sitting down while carrying an object (see Fig.~\ref{fig:teaser}). To address this issue, we present TokenHSI, a single, unified transformer-based policy capable of multi-skill unification and flexible adaptation. The key insight is to model the humanoid proprioception as a separate shared token and combine it with distinct task tokens via a masking mechanism. Such a unified policy enables effective knowledge sharing across skills, thereby facilitating the multi-task training. Moreover, our policy architecture supports variable length inputs, enabling flexible adaptation of learned skills to new scenarios. By training additional task tokenizers, we can not only modify the geometries of interaction targets but also coordinate multiple skills to address complex tasks. The experiments demonstrate that our approach can significantly improve versatility, adaptability, and extensibility in various HSI tasks.
\end{abstract}
    
\section{Introduction}
\label{sec:intro}

Generating diverse and life-like Human-Scene Interactions (HSI) through physics simulation is a fascinating yet challenging task for computer animation and embodied AI. In real-world scenarios, humans function as general-purpose agents, proficient in executing a wide variety of complex interaction tasks and adept at adapting to novel contexts. Driven by this, we aim to establish a unified controller for learning versatile human-scene interaction skills and to explore an effective method for adapting learned skills to new tasks and environments, thereby narrowing the gap between simulated characters and real-world humans.

Beyond methods~\cite{pan2024synthesizing, hassan2023synthesizing, xie2023hierarchical, merel2020catch, chao2021learning, xu2024humanvla, wang2024sims, gao2024coohoi} that develop controllers solely focused on a single interaction task, some advancements~\cite{xiao2023unified, tessler2024maskedmimic} have aimed to devise unified controllers with diverse skills. However, they have notable limitations in two main aspects: (1) Their controllers are primarily designed for interactions in static scenes, such as sitting or touching immovable objects, and therefore cannot be applied to dynamic scenarios that require manipulation skills, such as carrying those objects. This incomplete skill set results in limited applicability and diminishes the potential to accomplish tasks that require various synergies between diverse skills, such as composite tasks (\eg, sitting down while carrying an object) and long-term manipulation tasks by sequencing multiple skills. (2) These approaches focus more on in-domain settings and suffer from limited generalization capability in novel scenes,
overlooking the adaptation of learned skills to novel scenarios. The process of directly fine-tuning pre-trained policies for new tasks is inefficient~\cite{xu2023adaptnet}, thereby further constraining the adaptability of their controllers.

To address these challenges, we present \ours, the first physics-based character control framework designed for unifying diverse HSI skills within a single network while maintaining flexible adaptability to novel scenarios. \ours constructs a separate observation space by individually tokenizing the humanoid proprioception and multiple task states. During inference, the character is directed to perform a specific task by combining the proprioception token with the corresponding task token using the masking mechanism in the transformer encoder~\cite{vaswani2017attention}. That is, the proprioception token is shared across tasks. This design choice not only enables the unified learning of multiple HSI skills but also encourages motor knowledge sharing to boost performance. Through multi-task training, our proprioception tokenizer can generalize to a wide range of character states. To demonstrate the effectiveness, we train \ours to simultaneously learn four representative HSI skills using a single transformer network, including following, sitting, climbing, and carrying.

Transcending multi-skill unification, \ours further excels at quickly adapting its learned skills to tackle new, more challenging HSI tasks. Given that our transformer policy enables variable length inputs, we can introduce additional task tokenizers to adapt the pre-trained policy to new tasks and environments both robustly and efficiently. This is thanks to the effective generalization of our proprioception tokenizer which is trained across diverse tasks, and the reuse of prior task tokenizers relevant to new contexts via the masking mechanism. 
Although the concurrent works~\cite{tessler2024maskedmimic,he2024hover} also utilize the transformer architecture with a shared proprioception tokenizer to develop a single controller, \ours differs from these approaches since we not only unify diverse HSI tasks but also further unleash the model's flexible and efficient adaptability to new tasks. Our framework facilitates the adaptation to novel task configurations by training only few additional parameters once the foundational skills are acquired, including task tokenizers, as well as adapter layers for the multilayer perceptron (MLP) based action head that predicts the actions. It eliminates the need to fine-tune the full parameters of the pre-trained policy, thereby enhancing the efficiency of the adaptation process. \ours demonstrates significantly improved sample efficiency and performance compared to recent policy adaptation methods~\cite{xu2023adaptnet,xu2023composite}.

We conduct extensive experiments on a variety of HSI tasks, including skill composition, object and terrain shape variation, and long-horizon task completion in complex environments; See an overview in Fig.~\ref{fig:teaser}. Our results demonstrate that \ours, despite its simplicity and efficiency, significantly outperforms existing methods on these challenging HSI benchmarks. Our contributions can be summarized as follows:

\begin{enumerate}
\item We present \ours, a novel transformer-based physical character controller that integrates versatile HSI skills within a single, unified policy.

\item Once trained, our approach enables flexible and efficient policy adaptation to novel HSI tasks, avoiding the full fine-tuning of the pre-trained policy.

\item We propose a dedicated tokenizer to encode proprioception, effectively facilitating both multi-task training and policy adaptation.

\end{enumerate}

\section{Related Work}
\label{sec: related_work}

\subsection{Human-Scene Interaction}
Recent advancements in human-scene interaction motion synthesis can be categorized into two main formulations: data-driven kinematic generation and physics-based character control. Traditional approaches have synthesized human-object interactions, such as grasping, using inverse kinematics~\cite{huang1995multi, elkoura2003handrix, kim2000neural, koga1994planning, aydin1999database} and optimal control~\cite{jain2011controlling}. Recently, a significant body of work has introduced data-driven methods~\cite{lee2018interactive, zhang2021manipnet, xu2023interdiff, li2023object, starke2019neural, xu2024interdreamer,li2024controllable,lee2023locomotion,zhang2024roam,pi2023hierarchical,zhang2022couch,jiang2024autonomous,lu2024choice,diller2024cg,jiang2024scaling,huang2023diffusion,zhao2023synthesizing,zhang2022wanderings,yi2024generating,cong2024laserhuman,cen2024generating,wang2021scene,wang2021synthesizing} that utilize large-scale datasets~\cite{hassan2021stochastic,li2023object,mahmood2019amass,fu20213d,bhatnagar2022behave,xu2025interact,liu2024core4d,xie2024intertrack} and advanced generative motion models~\cite{tevet2023human,ling2020character,chen2024taming,shi2024interactive,holden2020learned,holden2017phase,xiao2025motionstreamer,lu2024scamo,dai2024motionlcm,lu2023humantomato,lin2023motion,fan2024freemotion,zhang2024force,motionlcm-v2,zhou2024emdm,wan2024tlcontrol,xie2024omnicontrol,li2024lodge}, to synthesize high-quality human-scene interaction motions. However, these methods often overlook the importance of the physical plausibility.

To address this, traditional physics-based methods utilize hand-crafted controllers to implement physics-based grasping skills~\cite{zhao2013robust, pollard2005physically}. For instance, \cite{zhao2013robust} constructed a hand-grasping database to generate grasping poses for PD controllers, enabling real-time physics-based hand-grasping synthesis. As the complexity of human-object interactions increases, trajectory optimization offers a versatile and effective approach for synthesizing complex skills~\cite{liu2008synthesis, liu2009dextrous, ye2012synthesis, mordatch2012contact, mordatch2012discovery,chen2025hifar,liu2010sampling,liu2015improving}. However, these optimization-based methods are typically offline and cannot be deployed in real-time applications. More recently, deep reinforcement learning (DRL) has emerged as a powerful framework for learning motor skills in physically simulated agents~\cite{peng2017deeploco, peng2018deepmimic, bergamin2019drecon, park2019learning, tan2018sim, peng2021amp, hassan2023synthesizing, liu2022motor, clegg2018learning, peng2022ase, dou2023c, xu2025intermimic,cui2024anyskill,zhu2023neural,yao2024moconvq,serifi2024vmp} and real-world humanoid robots~\cite{cheng2024expressive,ji2024exbody2,fu2024humanplus,he2024hover,heomnih2o,he2025asap,huang2025learning,wang2025beamdojo,zhang2024wococo}. This approach has successfully synthesized numerous impressive human-object interaction skills that were previously unattainable, including basketball dribbling~\cite{liu2018learning,wang2024skillmimic}, skateboarding~\cite{liu2017learning}, playing tennis~\cite{zhang2023vid2player3d,wang2024strategy}, solving a Rubik's Cube with hands~\cite{akkaya2019solving}, using chopsticks~\cite{yang2022learning}, and interacting with everyday objects~\cite{hassan2023synthesizing, xiao2023unified, pan2024synthesizing,li2024physics,wang2024sims,gao2024coohoi,li2025learning,liu2024mimicking,braun2024physically,luo2024omnigrasp,wu2024human}. However, despite these significant advancements, existing methods still fall short of establishing a single, comprehensive controller for integrating diverse skills, such as climbing, contacting, and manipulation.

\subsection{Unified Character Controller}
Beyond controllers designed for limited skills~\cite{peng2018deepmimic, hassan2023synthesizing, liu2017learning}, recent works aim to develop unified controllers for a broader range of skills. Notably, studies~\cite{luo2023perpetual, won2019learning, won2020scalable} have enhanced the capabilities of physics-aware motion imitation, enabling support for large-scale reference motions and accommodating a variety of body shapes. To improve the flexibility of learned skills, \cite{peng2022ase, won2022physics, luo2024universal, dou2023c, ControlVAE, yao2024moconvq, serifi2024vmp} have established motion manifolds using physics simulations to model versatile and reusable skills while also training additional controllers for specific tasks. UniHSI~\cite{xiao2023unified} proposes a unified controller for various contact-based human-scene interaction tasks, utilizing automatic task generation through large language models (LLMs). More recently, \cite{tessler2024maskedmimic, wang2024pacer+,he2024hover} introduced a masking mechanism to train controllers that follow on-demand tracking targets. In contrast, our approach not only incorporates the masking mechanism to unify various HSI tasks across different categories—such as following, sitting, climbing, and carrying—but also explores effective methods to adapt the skills learned by our unified character controller to new tasks and environments.

\section{Methodology}
\label{sec:method}

\begin{figure*}[t]
  \centering
   \includegraphics[width=0.75\linewidth]{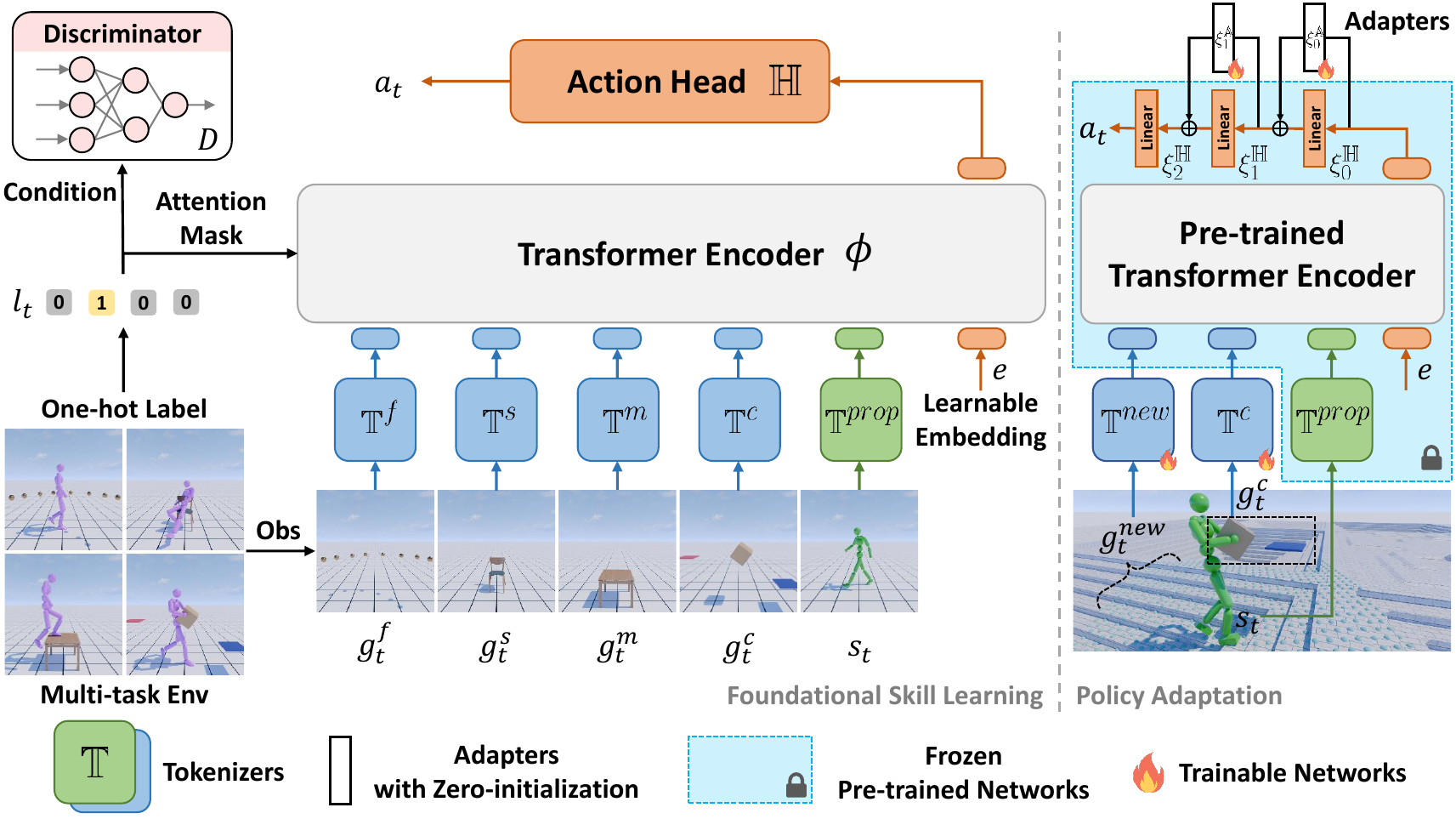}
   \caption{\ours consists of two stages: (left) foundational skill learning and (right) policy adaptation. Through multi-task policy training, the proposed framework learns versatile interaction skills in a single transformer network. Theses learned skills can be flexibly adapted to more challenging HSI tasks by training the lightweight modules, \eg, $ \mathbb{T}^{new}$, $\mathbb{T}^{c}$, and $ \xi^{\mathbb{A}}= \{ \xi^{\mathbb{A}}_0, \xi^{\mathbb{A}}_1 \}$.}
   \label{fig:pipeline}
\end{figure*}

\subsection{Overview} \label{sec:overview}

We focus on developing a physical HSI controller capable of unifying diverse interaction skills within a single network. Beyond versatility, the unified controller should be able to generalize learned skills to novel settings, enabling it to tackle more complex HSI tasks. Driven by these goals, we propose a transformer-based policy network, leveraging its support for variable length inputs to seamlessly incorporate an arbitrary number of tasks.

During foundational skill learning (Fig.~\ref{fig:pipeline} left), we standardize the observation lengths of different tasks using separate task tokenizers $\mathbb{T}^{task}$. Importantly, we adopt a proprioception tokenizer $\mathbb{T}^{prop}$ dedicated to processing the character state $s_t$, in contrast to existing methods~\cite{hassan2023synthesizing,pan2024synthesizing,xiao2023unified,rempe2023trace,gao2024coohoi,xu2024humanvla,wang2024sims} that rely on a joint character-goal state space. When training a specific task, we utilize the masking mechanism within the transformer encoder $\phi$~\cite{vaswani2017attention} to combine the proprioception token with the relevant task token. Using the shared $\phi$ and $\mathbb{T}^{prop}$ can effectively encourage motor knowledge sharing across tasks, thereby improving the multi-task training performance.

Once trained, we adapt the unified policy to novel tasks by reusing prior tokenizers { $\mathbb{T}^{prop}$, $\mathbb{T}^{task}$ } and adding additional task tokenizers $\mathbb{T}^{new}$ if new task observations exist. To facilitate the adaptation, we integrate zero-initialized adapter layers $\xi^{\mathbb{A}}$~\cite{hu2022lora} into the action head module $\mathbb{H}$. This transformer-based policy adaptation is efficient as the proprioception tokenizer $\mathbb{T}^{prop}$ trained across diverse tasks gains effective generalization to a wide range of character states.  We achieve several challenging HSI tasks through the adaptation of foundation skills, including skill composition, object/terrain shape variation, and long-horizon task completion.

Sec.~\ref{subsec: Physics-aware character controller} provides the formulation for physics-based character control. Sec.~\ref{subsec: Fundational Skill Learning} details the construction and training of the unified transformer policy for simultaneously learning multiple foundational HSI skills. Sec.~\ref{subsec: Skill Adaptation} introduces our novel policy adaptation technique based on the flexible transformer architecture.

\subsection{Physics-based Character Control}~\label{subsec: Physics-aware character controller}

\noindent \textbf{Formulation.}
We use goal-conditioned reinforcement learning to formulate character control as a Markov Decision Process (MDP) defined by states, actions, transition dynamics, a reward function $r$, and a discount factor $\gamma$. The reward $r_t \in \mathcal{R}$ is calculated by a style reward $r_t^{style}$~\cite{peng2021amp} and a task reward $r_t^{task}$. The policy aims to maximize the accumulated discounted reward $\sum_{t=0}^T \gamma^t r_t$. We use the widely adopted Proximal Policy Optimization (PPO) algorithm \cite{schulman2017proximal} to train the control policy model.

\noindent \textbf{State and Actions.} The state $o_t \triangleq (s_t, g_t)$ consists of the character state $s_t$ and the goal state $g_t$, which is recorded in the character's local coordinate frame. We use PD controllers at each degree of freedom of the humanoid. The action $a_t$ specifies joint target rotations for the PD controllers, resulting in a $32$-dim action space.


\subsection{Foundational Skill Learning}~\label{subsec: Fundational Skill Learning}

\noindent \textbf{Policy Architecture.} We propose a scalable architecture that supports variable length inputs to enable multi-task training. As illustrated in Fig.~\ref{fig:pipeline} (left), our policy consists of four key parts:
\textit{(1) Multiple tokenizers.} Each tokenizer $(\mathbb{T}^f, \mathbb{T}^s, \mathbb{T}^m, \mathbb{T}^c$, or $\mathbb{T}^{prop})$ standardizes its corresponding observation $(g_t^{f}, g_t^{s}, g_t^{m}, g_t^{c}$, or $s_t)$ into a $64$-dim feature. The tokenizers are implemented as separate MLPs with $3$ hidden layers of $[256, 128, 64]$ units. \textit{(2) A transformer encoder} $\phi$ comprises $4$ encoder layers~\cite{vaswani2017attention}, each containing $2$ attention heads and a $512$-dim feed-forward layer. It fuses all information through self-attention mechanism. \textit{(3) A 64-dim learnable embedding} $e$ functions as the output token. \textit{(4) An action head} $\mathbb{H}=\{ \xi_0^\mathbb{H}, \xi_1^\mathbb{H} , \xi_2^\mathbb{H}  \}$ produces the final action $a_t$. It is also modeled by an MLP network, but with $3$ hidden layers of $[1024, 512, 32]$ units.

\noindent \textbf{Foundational Skills.} We consider four types of HSI tasks. Their task observations are formulated as: (1) $g_t^{f} \in \mathbb{R}^{2 \times 10}$, the following task~\cite{rempe2023trace,wang2024pacer+,hallak2024plamo}, tracking a target path described by ten 2D coordinates; (2) $g_t^{s} \in \mathbb{R}^{38}$, the sitting task~\cite{hassan2023synthesizing,pan2024synthesizing,xiao2023unified}, approaching and sitting on a target object; (3) $g_t^{m} \in \mathbb{R}^{27}$, the climbing task~\cite{khoshsiyar2024partwisempc}, approaching and climbing onto a target object; (4) $g_t^{c} \in \mathbb{R}^{42}$, the carrying task~\cite{hassan2023synthesizing,gao2024coohoi}, moving a box from its initial location to a target location. See Sec.~\ref{sec:supp_task_foundational} of the supplementary for the detailed task designs.


\noindent \textbf{Multi-task Policy Training.} We train the transformer policy in a multi-task environment with flat ground. During resetting, the environment is assigned to a specific task based on a pre-defined probability distribution: $10\%$ for the path-following task and $30\%$ each for the other three tasks. The assigned task's reward function is then applied to compute the task reward $r_t^{task}$. We encode the active task using a one-hot label $l_t$. The training objective is to obtain a multi-task policy $\pi(a_t|e, s_t,g^f_t,g^s_t,g^m_t,g^c_t)$. Leveraging the current task label $l_t$ as an attention mask, we suppress features from non-target tasks in the self-attention mechanism. Critically, the character state $s_t$ and the learnable embedding $e$ persistently participate in all attention computations. Thus, the multi-task policy reduces to a single-task formulation $\pi(a_t|e,s_t,g_t^{l_t})$. To optimize the policy, we employ a value function $V(s_t,g^f_t,g^s_t,g^m_t,g^c_t)$, modeled by an MLP network with 4 hidden layers of $[2048, 1024, 512, 1]$ units. The irrelevant inputs will be padded to zeros, reducing the value function to $V(s_t,g_t^{l_t})$. Besides, we condition the motion discriminator $D$ on the one-hot label $l_t$ to prevent the policy from learning motor skills unrelated to the current task. For instance, without the conditional $D$, the character may incorrectly produce squatting motions during path-following. The discriminator is modeled by a three layer MLP with $[1024, 512, 1]$ units. We run numerous environments in parallel~\cite{makoviychuk2021isaac} to achieve large-scale training.

\subsection{Transformer-based Policy Adaptation}~\label{subsec: Skill Adaptation}

In this section, we demonstrate how to adapt the pre-trained policy to address diverse new HSI tasks. We validate our approach on three challenging tasks. As depicted in Fig.~\ref{fig:pipeline} (right), the adaptation process first freezes these components: $\mathbb{T}^{prop}$, $e$, $\mathbb{H}$, and $\phi$ to ensure that the adapted model remains effective for prior skills. Then, we introduce zero-initialized adapter layers $ \xi^{\mathbb{A}}= \{ \xi^{\mathbb{A}}_0, \xi^{\mathbb{A}}_1 \}$ to the action head $\mathbb{H}$ to enhance the adaptation efficiency. We train a new motion discriminator $D$ from scratch during each task. Next, we detail how we reuse pre-trained task tokenizers and introduce new tokenizers to learn these tasks. Detailed task designs are provided in Sec.~\ref{sec:supp_task_advanced} in the supplementary.


\noindent \textbf{Skill Composition.} We aim to create composite motions, such as sitting down while carrying a box (see Fig.~\ref{fig:result} (b)). We reuse and \textit{freeze} the task tokenizers for the sitting $\mathbb{T}^s$ and the carrying $\mathbb{T}^c$ tasks. Furthermore, we introduce an additional trainable task tokenizer $\mathbb{T}^{new}$ to perceive the new task state $g_t^{new}$, which is designed to contain the states of the target object and the box. During the training, the model gradually learns to coordinate the two learned skills to complete the composite task. We also apply this method to combine other learned skills with the carrying skill, respectively.

\noindent \textbf{Object Shape Variation.} Possessing the ability to interact with a variety of objects is an essential feature of an HSI controller. The pre-trained policy learns a box-carrying skill. Therefore, we aim to adapt it to new environments in which the boxes are replaced with irregular objects, such as chairs and tables. Since we can use the same observation information, we directly fine-tune the pre-trained task tokenizer $\mathbb{T}^{c}$. After training, we obtain two additional fine-tuned tokenizers $\mathbb{T}^{c}_{chair}$ and $\mathbb{T}^{c}_{table}$, enabling the unified model to generalize to more diverse object categories.

\noindent \textbf{Terrain Shape Variation.} The basic skills of our policy are trained on flat ground. However, there is usually complex terrain in the real world, such as stairs. As illustrated in Fig.~\ref{fig:pipeline} (right), we generalize the trajectory following and carrying skills to stairs environments. Similarly, we directly fine-tune the prior task tokenizer $\mathbb{T}^f$ or $\mathbb{T}^c$. We further introduce an additional trainable task tokenizer $\mathbb{T}^{new}$ to encode the height map~\cite{rempe2023trace}.

\noindent \textbf{Long-horizon Task Completion.} Performing long-term tasks in complex environments challenges control models in achieving seamless skill transitions and collision avoidance. Existing methods~\cite{clegg2018learning,leeadversarial} tackle this problem by jointly fine-tuning multiple policies to achieve fluent transitions within the long skill chain. In contrast, our policy adaptation allows users to fine-tune only the lightweight task tokenizers. Additionally, we can effortlessly make the policy environment-aware by introducing a height tokenizer, which also facilitates the character's ability to avoid environmental obstacles. During inference, we employ a Finite State Machine (FSM) similar to~\cite{pan2024synthesizing} to enable automated task switching. Given a one-hot task label $l_t$ queried from the FSM, we use it as an attention mask to activate the corresponding task token and disable other tokens.

\section{Experiments}
\label{sec:experiments}

We conduct extensive experiments to evaluate foundational skill learning and policy adaptation. In Sec.~\ref{exp:foundation}, we evaluate the robustness of $4$ basic skills learned by our unified policy. In Sec.~\ref{exp:adaptation}, we demonstrate the efficiency advantage of our approach in adapting to complex HSI tasks. In Sec.~\ref{sec:supp_extensibiliy} of the supplementary, we further demonstrate our approach's extensibility through introducing novel out-of-domain skills.


\subsection{Evaluation on Foundational Skill Learning} \label{exp:foundation}

\noindent \textbf{Experimental Setup.} We compare our unified multi-task policy with policies trained individually for each task. The AMP~\cite{peng2021amp} framework is employed to train these specialists. Each policy is modeled by a four layer MLP with $[2048, 1024, 512, 32]$ units. All policies are trained with $4, 096$ parallel environments and $50k$ PPO~\cite{schulman2017proximal} iteration steps. $512$ trials are collected to calculate the success rate. We also measure the average error for all successful trials.

\textit{Follow.} We declare path-following to be successful if the pelvis is within $30$ cm (XY-planar distance) of the path endpoint. The error of a trial is calculated by $\frac{1}{n}\sum\limits_{t=1}^{n} \left \| x_t^{pelvis} - x_t^{tar} \right \|_2$, where $n$ is the episode length, and $x_t^{tar}$ is the dynamically sampled target pelvis position. Both training and testing paths are procedurally generated~\cite{rempe2023trace}.

\textit{Sit and Climb.} The success criteria for both tasks require the pelvis to lie within a $20$ cm radius sphere centered at the target location. Error is the minimal 3D distance between the pelvis and its target location throughout the trial. The object is spawned between $1$ m and $5$ m from the character with randomized orientation. The sitting task uses $49$ training and $26$ testing objects, while the climbing task utilizes $38$ training and $26$ testing objects. 

\textit{Carry.} The success criterion and error calculation are the same as above, with the reference point altered from the character's pelvis to the box's centroid. The box is placed between $1$--$9$ m from the character, with randomized orientation and height. The target location is uniformly sampled from a $10$ m $\times$ $10$ m 2D horizontal (XY) plane. $9$ boxes with different sizes are used for training and $9$ for testing.

\noindent \textbf{Results.} Tab.~\ref{tab:basic} presents the quantitative results, indicating that compared to specialists, \ours achieves higher success rates while maintaining comparable errors on all foundational skills. Our unified policy exhibits superior cross-task generalization performance, making it a versatile and flexible alternative to individual controllers.

\noindent \textbf{Ablation on Shared $\mathbb{T}^{prop}$.} We answer the question: can using a shared proprioception tokenizer $\mathbb{T}^{prop}$ really boost the performance? We design a variant of our approach, namely Ours (w/o $\mathbb{T}^{prop}$), where the $\mathbb{T}^{prop}$ is discarded and the proprioception $s_t$ is added into the input of each task tokenizer. We observe a general decrease in the success rate across all tasks, suggesting that using a shared  $\mathbb{T}^{prop}$ indeed contributes to effective knowledge sharing across tasks.

\subsection{Evaluation on Policy Adaptation} \label{exp:adaptation}
We evaluate the policy adaptation on more challenging HSI tasks. Comparative experiments with existing methods demonstrate the high efficiency of our approach. \ours 
achieves efficient policy adaptation through a simple and unified design, avoiding the need to train new policies from scratch, fully fine-tune prior models, or implement sophisticated architectures like AdaptNet~\cite{xu2023adaptnet} and CML~\cite{xu2023composite}. 

\begin{table}[t]
    \centering
    \resizebox{0.65\linewidth}{!}{
        \begin{tabular}{lc|cc}
            \textbf{Task} & \textbf{Method} & \textbf{Success Rate (\%)} & \textbf{Error (cm)} \\ \hline\hline
            \multirow{3}{*}{{Follow}}       & \small{Specialist}     & 98.7$\pm$0.5  & \cellcolor{grey!17}\textbf{6.5$\pm$0.0} \\
                                            & \small{Ours (w/o $\mathbb{T}^{prop}$)}           & 99.3$\pm$0.3  & 9.7$\pm$0.2 \\
                                            & \small{Ours}           & \cellcolor{grey!17}\textbf{99.7$\pm$0.0}  & 9.3$\pm$0.1 \\ \hline
            \multirow{3}{*}{{Sit}}          & \small{Specialist}     & 98.2$\pm$2.0  & \cellcolor{grey!17}\textbf{5.6$\pm$0.0}  \\
                                            & \small{Ours (w/o $\mathbb{T}^{prop}$)}           & 98.7$\pm$0.4  & 5.6$\pm$0.1 \\
                                            & \small{Ours}           & \cellcolor{grey!17}\textbf{99.6$\pm$0.2}  & 5.6$\pm$0.2  \\ \hline
            \multirow{3}{*}{{Climb}}        & \small{Specialist}     & 99.7$\pm$0.1  & \cellcolor{grey!17}\textbf{2.4$\pm$0.2}  \\
                                            & \small{Ours (w/o $\mathbb{T}^{prop}$)}           & 99.5$\pm$0.2  & 3.1$\pm$0.8 \\
                                            & \small{Ours}           & \cellcolor{grey!17}\textbf{99.8$\pm$0.1}  & 2.7$\pm$0.3  \\ 
                                            \hline
            \multirow{3}{*}{{Carry}}        & \small{Specialist}     & 83.1$\pm$5.0  & 5.1$\pm$0.2  \\
                                            & \small{Ours (w/o $\mathbb{T}^{prop}$)}           & 90.9$\pm$3.3  & 6.0$\pm$0.5 \\
                                            & \small{Ours}           & \cellcolor{grey!17}\textbf{92.2$\pm$6.7}  & \cellcolor{grey!17}\textbf{4.2$\pm$0.6} 
        \end{tabular}
    }
    \caption{Quantitative comparison between our unified multi-task policy and specialist policies across four foundational HSI skills. Values are reported in the format of mean$\pm$std.}
    \label{tab:basic}
\end{table}

\subsubsection{Skill Composition} \label{exp:comp}
\noindent \textbf{Experimental Setup.} We consider three possible combinations: \textit{Follow + Carry}, \textit{Sit + Carry}, \textit{Climb + Carry}. For every composite task, the character should perform the core interaction task (\eg, sitting) while continuously carrying the box. Task success requires simultaneous fulfillment of: (1) core task completion; (2) box-carrying maintenance (carrying is deemed failed if the box's lowest point is less than $20$ cm from the ground). Please refer to Sec.~\ref{exp:foundation} for the success criterion, error calculation, object dataset, and object initialization for each core task. The character is initialized in a standing pose while holding a box, which means the walk-to-pick-up stage is omitted.

\noindent \textbf{Baselines.} We first implement Scratch, which trains policies from scratch via AMP~\cite{peng2021amp}, to benchmark the difficulty of different tasks. Then, we adopt CML~\cite{xu2023composite}, the current SOTA approach for composite motion learning. Specifically, we apply the incremental learning scheme to adapt each core task's specialist policy. Given that our approach uses two skill priors and CML only uses one, we further establish CML (dual) that can jointly coordinate two pre-trained policies (one for the core task and another for the carrying task) to ensure a fair comparison. See Sec.~\textcolor{cvprblue}{C} of the supplementary for the details of CML (dual). All base policies are pre-trained with $50k$ iterations in Sec.~\ref{exp:foundation}, except the box-carrying policy, which is further trained to $100k$ iterations. This eliminates the performance gap between the carrying specialist ($95.7\%$) and \ours ($97.4\%$).  The training on skill composition exploits $4, 096$ environments and $5k$ iterations.


\begin{table}[t]
    \centering
    \resizebox{0.95\linewidth}{!}{
        \begin{tabular}{l|cc|cc|cc}
            \multirow{2}{*}{{\textbf{Method}}} & \multicolumn{2}{c|}{\textbf{Follow + Carry}} & \multicolumn{2}{c|}{\textbf{Sit + Carry}} & \multicolumn{2}{c}{\textbf{Climb + Carry}} \\ 
            & \textbf{\small{Succ. (\%)}} & \textbf{\small{Err. (cm)}} & \textbf{\small{Succ. (\%)}} & \textbf{\small{Err. (cm)}} & \textbf{\small{Succ. (\%)}} & \textbf{\small{Err. (cm)}} \\ \hline\hline
            Scratch~\cite{peng2021amp} & 98.1$\pm$0.7 & 12.5$\pm$0.1 & 82.9$\pm$5.0 & 5.9$\pm$0.1 & 26.8$\pm$37.8 & 5.7$\pm$4.8 \\
            CML~\cite{xu2023composite} & 97.6$\pm$0.8 & \cellcolor{grey!17}\textbf{9.9$\pm$0.0} & \cellcolor{grey!17}\textbf{96.7$\pm$0.6} & 6.2$\pm$0.2 & 68.3$\pm$21.9 & 4.8$\pm$0.5 \\
            CML (dual) & 99.1$\pm$0.2 & 10.0$\pm$0.0 & 94.9$\pm$2.0 & 5.9$\pm$0.0 & 51.3$\pm$40.2 & 9.1$\pm$7.3 \\ \hline
            Ours (w/o $\mathbb{T}^{prop}$) & \cellcolor{grey!17}\textbf{99.2$\pm$0.5} & 11.5$\pm$0.2 & 82.0$\pm$4.2 & 6.9$\pm$0.9 & 81.9$\pm$9.0 & 5.0$\pm$0.5 \\
            Ours & 99.1$\pm$0.4 & 10.8$\pm$0.3 & 96.3$\pm$1.1 & \cellcolor{grey!17}\textbf{5.5$\pm$0.0} & \cellcolor{grey!17}\textbf{99.2$\pm$0.1} & \cellcolor{grey!17}\textbf{3.5$\pm$0.1} \\
        \end{tabular}
    }
    \caption{Quantitative results across skill composition tasks.}
    \label{tab:comp}
\end{table}

\begin{figure}[t]
  \centering
   \includegraphics[width=0.8\linewidth]{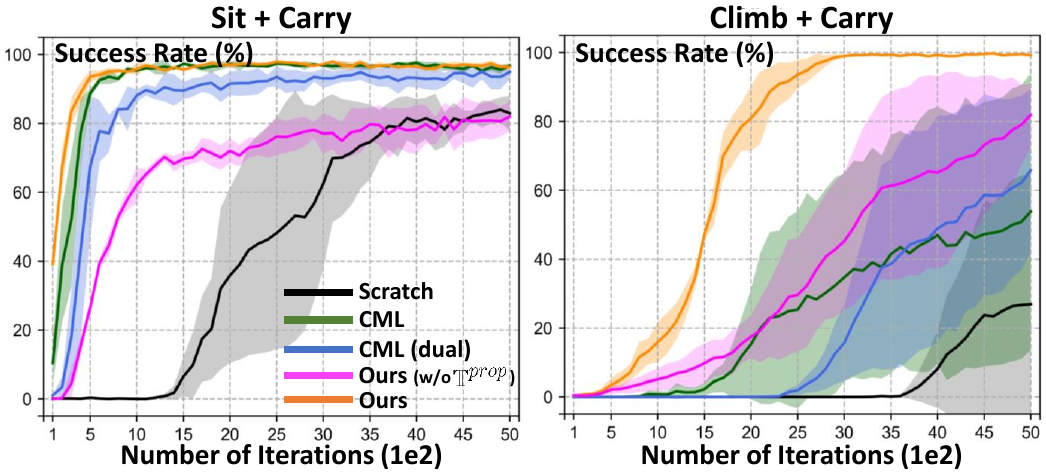}
   \caption{Learning curves comparing the efficiency on skill composition tasks using \ours, policies trained from scratch~\cite{peng2021amp}, CML~\cite{xu2023composite}, and its improved version CML (dual). Colored regions denote mean values $\pm$ a standard deviation based on $3$ models initialized with different random seeds.}
   \label{fig:comp}
\end{figure}

\begin{figure*}[t]
  \centering
   \includegraphics[width=0.75\linewidth]{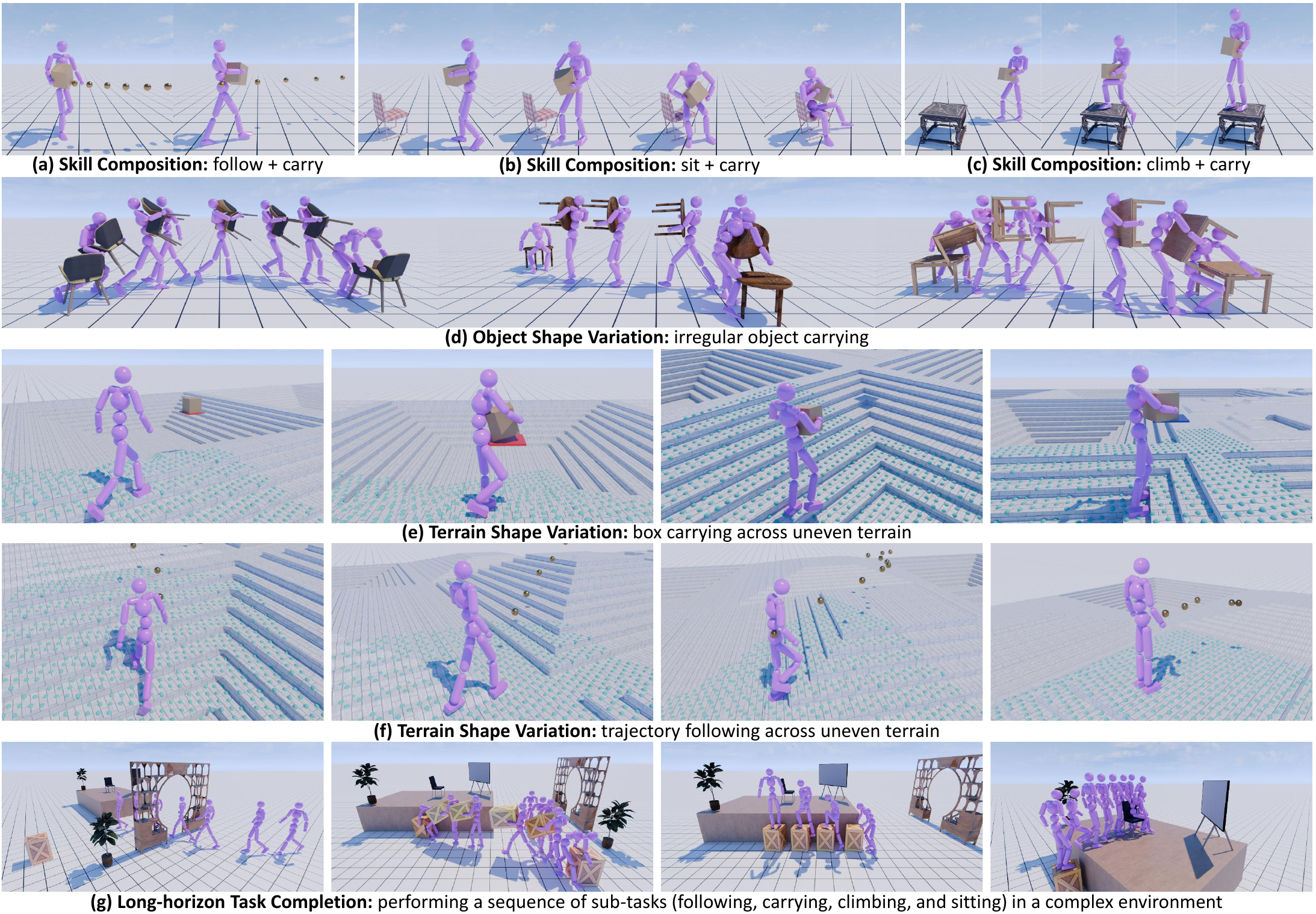}
   \caption{Through policy adaptation, \ours can generalize learned foundational skills to more challenging scene interaction tasks.}
   \label{fig:result}
\end{figure*}

\noindent \textbf{Results.} We report quantitative results in Tab.~\ref{tab:comp}. For the first two easier tasks--where even the Scratch attains $98.1\%$ and $82.9\%$ success rates--\ours achieves near-optimal performance. Despite the drastic performance degradation observed in baselines as task difficulty escalates (Scratch: $80\%+\rightarrow26.8\%$, CML: $90\%+\rightarrow68.3\%$, and CML (dual): $90\%+\rightarrow51.3\%$), our approach still maintains a high success rate of $99.2\%$ on the third challenging \textit{Climb + Carry} task. In Fig.~\ref{fig:comp}, we further show the learning curves of different methods to compare their convergence processes intuitively. \ours exhibits superior \textbf{efficiency and stability} compared to all baselines, especially on the \textit{Climb + Carry} task. We attribute these advantages to (1) the proprioception tokenizer $\mathbb{T}^{prop}$ trained across diverse tasks offers better generalization performance than the base policies (trained on a single task) used by CML and CML (dual); (2) the attention mechanism enables skill composition at an earlier stage (within the transformer encoder $\phi$'s latent feature space), whereas CML-based approaches are limited to post-composition in the action space. The synthesized composite motions are shown in Fig.~\ref{fig:result} (a-c).

\noindent \textbf{Ablation on Shared $\mathbb{T}^{prop}$.} We validate the importance of the shared $\mathbb{T}^{prop}$ on the policy adaptation. We adapt a base model trained without $\mathbb{T}^{prop}$ to learn composition tasks. During training, the additional $\mathbb{T}^{new}$ observes $(s_t, g_t^{new})$. The results show a significant performance (both robustness and efficiency) drop in the last two difficult tasks, which further demonstrates that the design choice of modeling the proprioception as a separate token is necessary.

\begin{table}[t]
    \centering
    \resizebox{0.7\linewidth}{!}{
        \begin{tabular}{lc|cc}
            \textbf{Object} & \textbf{Method} & \textbf{Success Rate (\%)} & \textbf{Error (cm)} \\ \hline\hline
            \multirow{3}{*}{{Chair}}     & \small{Finetune}   & 87.5$\pm$0.6 & 6.4$\pm$0.2 \\
                                            & \small{AdaptNet}~\cite{xu2023adaptnet}   & 84.5$\pm$3.0  & 6.8$\pm$0.5  \\
                                            & \small{Ours}   & \cellcolor{grey!17}\textbf{88.8$\pm$3.1} & \cellcolor{grey!17}\textbf{5.6$\pm$0.2} \\ \hline
            \multirow{3}{*}{{Table}}     & \small{Finetune}   & 83.4$\pm$1.6 & \cellcolor{grey!17}\textbf{6.0$\pm$0.1} \\
                                            & \small{AdaptNet}~\cite{xu2023adaptnet}   & 82.4$\pm$3.9 & 6.4$\pm$0.3 \\
                                            & \small{Ours}   & \cellcolor{grey!17}\textbf{83.6$\pm$1.6} & 6.3$\pm$0.2
         
        \end{tabular}
    }
    \caption{Quantitative results across object shape variation tasks.}
    \label{tab:obj}
\end{table}

\begin{figure}[t]
  \centering
   \includegraphics[width=0.8\linewidth]{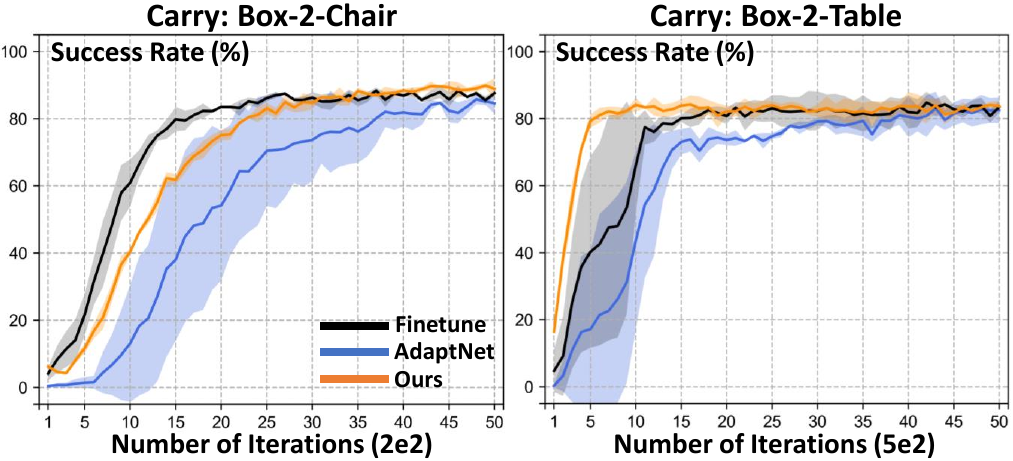}
   \caption{Learning curves comparing the efficiency on object shape variation tasks using \ours, full fine-tuning of pre-trained policies, and AdaptNet~\cite{xu2023adaptnet}.}
   \label{fig:obj}
\end{figure}

\subsubsection{Object Shape Variation} \label{exp:object_shape_variation}

\noindent \textbf{Experimental Setup.} We evaluate on two irregular object categories: \textit{Chair} and \textit{Table}, with the training-testing instance numbers being $63$--$27$ and $21$--$9$, respectively. The success criterion and error calculation are identical to the carrying task in Sec.~\ref{exp:foundation}. We add a constraint requiring the object's height axis $<30^\circ$ from the ground's vertical axis (Z) to filter out incorrect object placement poses. The object is placed on the ground between $1$--$5$ m from the character, with randomized orientation. The target location is still uniformly sampled from a $10$ m $\times$ $10$ m 2D area.

\noindent \textbf{Baselines.} We first establish Finetune, which fine-tunes a specialized box-carrying policy. Then, we employ AdaptNet~\cite{xu2023adaptnet}, the current SOTA network architecture for policy adaptation. The baselines' base box-carrying training requires $100k$ iterations, versus $50k$ for \ours, as detailed in Sec.~\ref{exp:comp}. The training on object shape variation adopts $4, 096$ environments and $10k$ iterations.

\noindent \textbf{Results.} Tab.~\ref{tab:obj} shows that our approach surpasses all baselines in success rate. As shown in Fig~\ref{fig:obj}, \ours exhibits efficiency and stability advantages across most comparisons, except when compared with Finetune on the \textit{Chair} category. Fine-tuning improves efficiency by overwriting all parameters, but it cannot retain prior skills. In contrast, our goal is skill-preserving adaptation. Under this requirement, \ours significantly outperforms the fair baseline AdaptNet~\cite{xu2023adaptnet}, validating our architectural superiority. The qualitative motions are presented in Fig.~\ref{fig:result} (d).

\subsubsection{Terrain Shape Variation} \label{exp:terrain_shape_variation}

\noindent \textbf{Experimental Setup.} We aim to adapt \textit{Follow} and \textit{Carry} skills trained on flat ground to uneven terrain, which necessitates that the policy adaptation be flexible enough to incorporate new observations (e.g., height perception). We consider four types of terrain blocks: stairs up, stairs down, obstacles, and flat ground, with an initialization probability of $[0.35, 0.35, 0.2, 0.1]$. We follow~\cite{rempe2023trace,wang2024pacer+} to calculate a walkable map for character initialization.

\noindent \textbf{Baselines.} We first train policies $\pi(a_t|s_t,g_t,h_t)$ from scratch, where $g_t$ is the corresponding task state and $h_t$ denotes the height map. We then employ AdaptNet~\cite{xu2023adaptnet} with the incorporation of the height map $h_t$ via its latent space injection. Similarly, we make \ours height-aware during adaptation through training a new task tokenizer $\mathbb{T}^{new}$ for processing $h_t$, as illustrated in Fig.~\ref{fig:pipeline}. The training on terrain shape variation uses $2, 048$ environments with $5k$ iterations for \textit{Follow} and $50k$ iterations for \textit{Carry}.

\begin{figure}[t]
  \centering
   \includegraphics[width=0.8\linewidth]{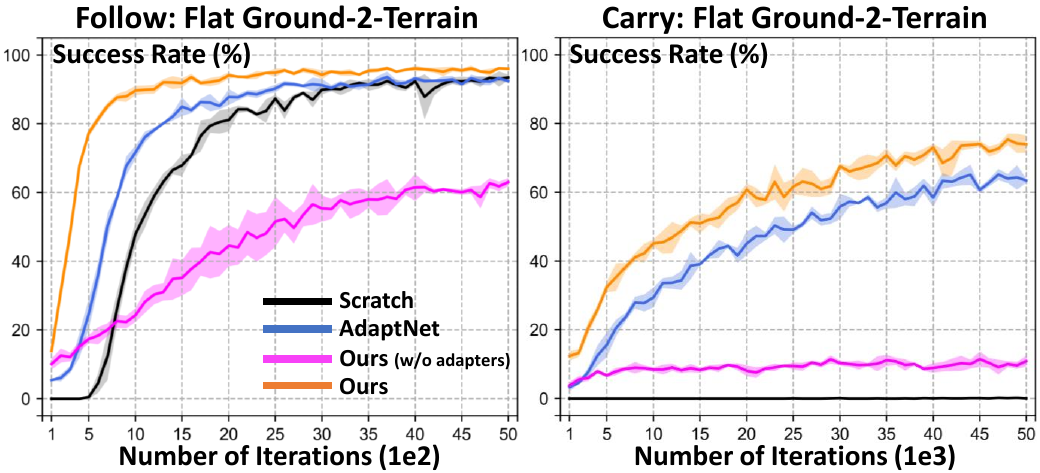}
   \caption{Learning curves comparing the efficiency on terrain shape variation tasks using \ours, Scratch~\cite{peng2021amp}, and AdaptNet~\cite{xu2023adaptnet}. We ablate the adapter layers during training.}
   \label{fig:terrain}
\end{figure}

\noindent \textbf{Results.} As indicated by Tab.~\ref{tab:comp_shape_terrain} and Fig.~\ref{fig:terrain}, our approach still maintains the efficiency advantage and outperforms all baselines in quantitative metrics. Compared to AdaptNet~\cite{xu2023adaptnet}, \ours supports both flexible input length and efficient learning, representing a significant advancement in policy adaptation for HSI. The generated animations are shown in Fig.~\ref{fig:result} (e) and (f).

\noindent \textbf{Ablation on Adapters.} During training, we remove the adapter layers $\xi^\mathbb{A}$ to conduct ablation studies. The results demonstrate that attaching learnable bypasses to the action head $\mathbb{H}$ is critical for improving performance.

\begin{table}[t]
    \centering
    \resizebox{0.8\linewidth}{!}{
        \begin{tabular}{l|cc|cc}
            \multirow{2}{*}{{\textbf{Method}}} & \multicolumn{2}{c|}{\textbf{Follow}} & \multicolumn{2}{c}{\textbf{Carry}} \\ & \textbf{\small{Succ. (\%)}} & \textbf{\small{Err. (cm)}}  & \textbf{\small{Succ. (\%)}} & \textbf{\small{Err. (cm)}}   \\ \hline\hline
            Scratch~\cite{peng2021amp} & 93.4$\pm$0.7 &  14.9$\pm$0.2 & 0 & --  \\
            AdaptNet~\cite{xu2023adaptnet} & 92.4$\pm$0.5 &  12.3$\pm$0.2 & 63.4$\pm$0.6 & 5.9$\pm$0.3 \\
            Ours (w/o adapters) & 63.0$\pm$1.1 & 22.5$\pm$0.4 & 10.8$\pm$1.5 & 7.7$\pm$0.1 \\
            Ours                       & \cellcolor{grey!17}\textbf{96.0$\pm$0.4} & \cellcolor{grey!17}\textbf{11.8$\pm$0.0} & \cellcolor{grey!17}\textbf{74.0$\pm$2.5} & \cellcolor{grey!17}\textbf{5.9$\pm$0.2} \\
        \end{tabular}
    }
    \caption{Quantitative results across terrain shape variation tasks.}
    \label{tab:comp_shape_terrain}
\end{table}

\begin{figure}[t]
  \centering
   \includegraphics[width=0.75\linewidth]{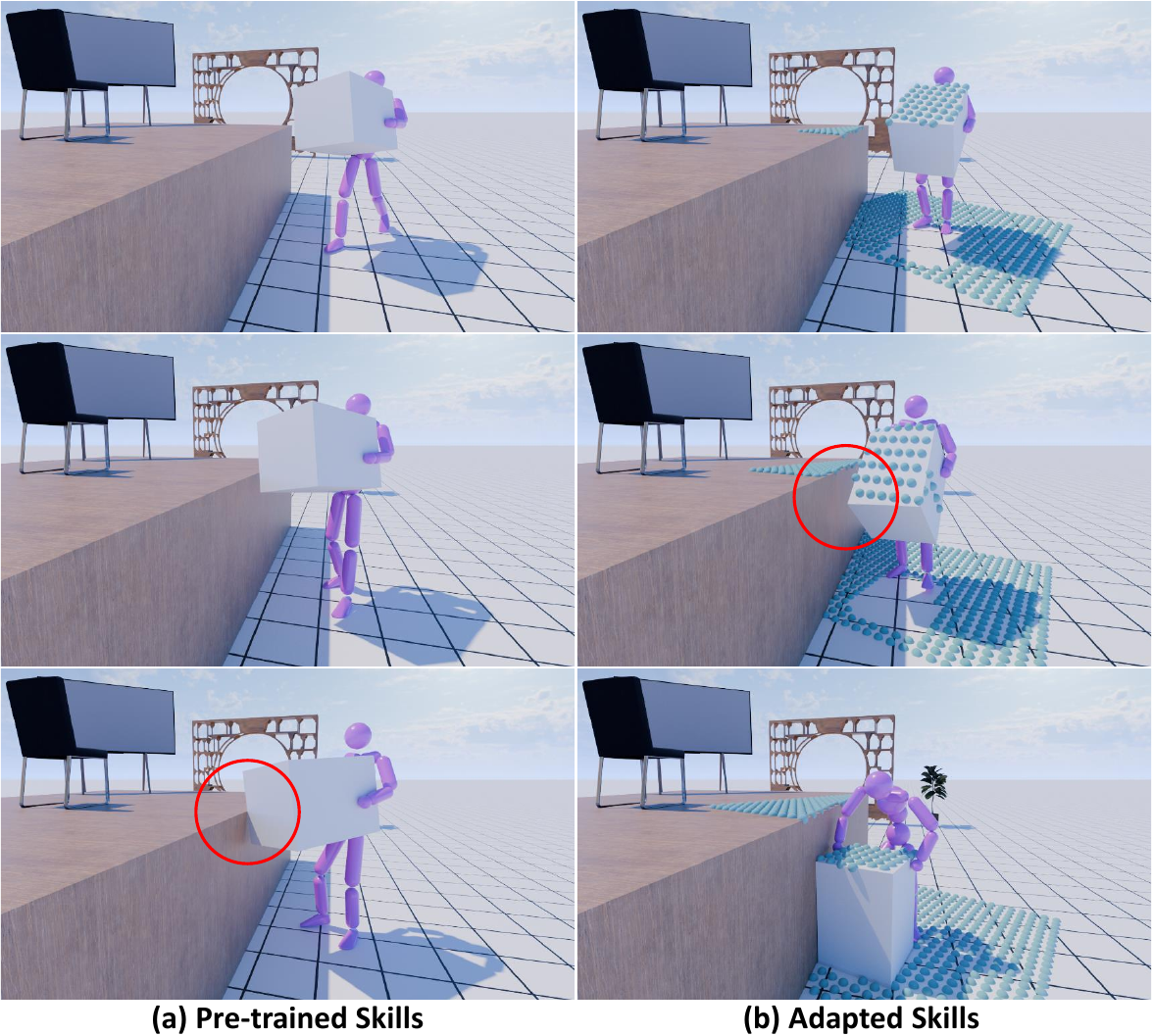}
   \caption{Long-horizon task completion by sequentially executing (a) pre-trained skills and (b) adapted skills by our approach.}
   \label{fig:longhorizon_bug}
\end{figure}

\subsubsection{Long-horizon Task Completion} \label{exp:long}
In this section, we validate how \ours facilitates long-horizon task completion in complex 3D scenes through policy adaptation. Directly executing multiple skills sequentially is prone to: (1) generating unseen transition states, which may cause subsequent skills to fail~\cite{clegg2018learning,leeadversarial}; (2) getting stuck by obstacles, since the application scenes are usually more cluttered than the training scenes, as illustrated in Fig.~\ref{fig:longhorizon_bug} (a). We design a long-horizon task containing these challenges. The task involves walking to pick up a box, placing it near a platform, and sitting on a chair on the platform after climbing onto it using the box. We first introduce a new height map tokenizer to make all skills environment-aware, and then iteratively fine-tune each task tokenizer in the complex scene. After policy adaptation, \ours successfully tackles the challenging long-horizon task, resulting in fluent skill execution and transition, as shown in Fig.~\ref{fig:result} (g). Fig.~\ref{fig:longhorizon_bug} (b) also demonstrates that our adapted skills can correctly place the box in preparation for the next climbing skill. \ours obviates the need to manually design transition states~\cite{xie2023hierarchical} and jointly fine-tune multiple policies~\cite{clegg2018learning,leeadversarial}. A more in-depth quantitative analysis is provided in Sec.~\ref{sec:supp_quan_long} of the supplementary.

\section{Discussion and Limitations}

In this work, we present TokenHSI to tackle the problem of unified synthesis of physical HSI animations. TokenHSI is a unified model that learns various HSI skills within a single transformer network and can flexibly generalize learned skills to novel tasks and environments through a simple yet efficient policy adaptation. We conduct extensive experiments to demonstrate that \ours significantly improves versatility, adaptability, and extensibility in HSI.

The main limitation is that learning these skills requires engineering of reward functions, which involve tedious trial-and-error processes. However, this is a general problem for the goal-oriented RL framework. In the future, we should explore effective approaches using human data~\cite{qiu2025humanoid} or internet knowledge~\cite{brohan2023rt} to reduce the cost on reward engineering. Besides, the current long-horizon task completion is still non-autonomous. A simulated humanoid that can complete complex, long-term tasks in realistic environments without human guidance remains an open problem.

\section{Acknowledgments}

We would like to thank Zhewen Zheng for his professional rendering techniques, which helped make appealing figures and videos in our paper. We also appreciate the anonymous reviewers for their constructive comments that improved the final version of this paper. This work is funded in part by the National Key R\&D Program of China (2022ZD0160201), HKU Startup Fund, and Shanghai Artificial Intelligence Laboratory. This work is partly supported by the Innovation and Technology Commission of the HKSAR Government under the ITSP-Platform grant (Ref: ITS/335/23FP) and the InnoHK initiative (TransGP project). The research work was in part conducted in the JC STEM Lab of Robotics for Soft Materials funded by The Hong Kong Jockey Club Charities Trust.

{
    \small
    \bibliographystyle{ieeenat_fullname}
    \bibliography{main}
}

\clearpage
\setcounter{page}{1}
\maketitlesupplementary

\setcounter{table}{0}
\renewcommand{\thetable}{\Alph{table}}
\renewcommand*{\theHtable}{\thetable}
\setcounter{figure}{0}
\renewcommand{\thefigure}{\Alph{figure}}
\renewcommand*{\theHfigure}{\thefigure}
\setcounter{section}{0}
\renewcommand{\thesection}{\Alph{section}}
\renewcommand*{\theHsection}{\thesection}

\section{Simulated Character}
\label{sec:supp_character}

\begin{figure}[t]
  \centering
   \includegraphics[width=1.0\linewidth]{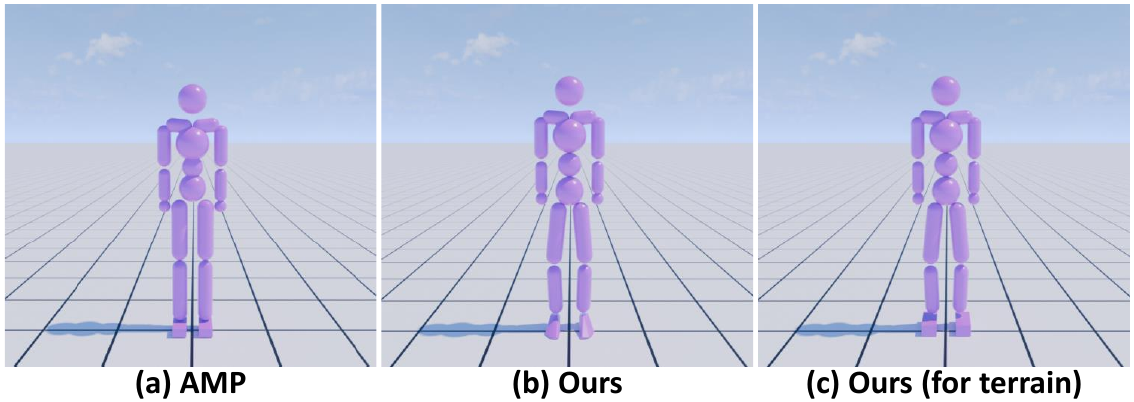}
   \caption{Different simulated character models. Building on (a) AMP's model, we devise two improved versions: (b) and (c), which are used for tasks on flat ground and  tasks on stairs terrain, respectively.}
   \label{fig:supp_character}
\end{figure}

\paragraph{Character Model Creation.} We apply a custom simulated character model, with $32$ degrees-of-freedom (DoF). This custom model is based on the character model used in AMP~\cite{peng2021amp}, which comprises $15$ rigid bodies, $12$ controllable joints, and $28$ DoF, as depicted in Fig.~\ref{fig:supp_character}~(a). While retaining most designs of AMP's character model, we introduce three improvements:

\begin{itemize}
\item (1) The $3$D relative positions of the lower body joints, including the hips, knees, and ankles, are adjusted to match those in the SMPL~\cite{loper2023smpl} human body model configured with a neutral gender and default shape parameters.
\item (2) The collision shapes of the foot rigid bodies are modified from rectangular boxes to realistic foot meshes using the method proposed by SimPoE~\cite{yuan2021simpoe}.
\item (3) The knee joints are upgraded from $1$-DoF revolute joints to $3$-DoF spherical joints. 
\end{itemize}
An illustration of our custom character model is available in Fig.~\ref{fig:supp_character}~(b). The primary motivation for building this custom model is two fold: the reference motion datasets are represented using SMPL parameters, and the kinematic structure of AMP's character model is different from that of SMPL. Consequently, directly copying rotation parameters to retarget these motions onto AMP's character model leads to unnatural lower body motions. In contrast, using our improved character model, which features a lower body structure consistent with SMPL, can significantly reduce retargeting errors and ensure more natural character motions. 


The designed simulated character is used for most tasks, except those involving stairs terrain, as illustrated in Fig.~\ref{fig:result} (e) and (f). This difference is due to inaccurate contact simulation between the meshed foot rigid bodies and the terrain in IsaacGym~\cite{makoviychuk2021isaac}. To address this issue, we revert the collision shapes of foot rigid bodies back to rectangular boxes, which are more simulation-friendly, as shown in Fig.~\ref{fig:supp_character}~(c).

\paragraph{The Proprioception $s_t$ and Action $a_t$.} The proprioception $s_t$ describes the simulated state of the character at each time step $t$. Following ASE~\cite{peng2022ase}, $s_t$ is constructed using a set of features, including the positions, rotations, linear velocities, and angular velocities of all rigid bodies. All features are expressed in the character's local coordinate frame, except for the root joint rotation, which is represented in the world coordinate frame. The $6$D rotation representation~\cite{zhou2019continuity} is employed. Notably, the root joint position is excluded from the proprioception. Combined, these features define a $222$D humanoid proprioception $s_t \in \mathbb{R}^{222}$. At each time step $t$, the control policy generates an action $a_t$, representing the target rotations for the PD controllers at each of the character's degrees-of-freedom. The target rotations for $3$D spherical joints are encoded using a $3$D exponential map~\cite{grassia1998practical}. Our character model has ten $3$-DoF spherical joints and two $1$-DoF revolute joints (\ie, left and right elbows), resulting in a $32$D action space $a_t \in \mathbb{R}^{32}$. No external forces are applied to any rigid body of the simulated character.

\section{Tasks}
\label{sec:supp_task}

In this section, we provide the implementation details about all tasks involved in this paper. Tab.~\ref{tab:supp_task_overview} presents an overview of all $12$ tasks, including $4$ foundational HSI tasks, $3$ skill composition tasks, $4$ object/terrain shape variation tasks, and $1$ long-horizon task. We begin by introducing the common settings shared across tasks in Section~\ref{sec:supp_task_pre}. Task-specific settings, such as task observations and reward functions, are detailed in the subsequent sections.

\begin{table*}[t]
    \centering
    \resizebox{1.0\linewidth}{!}{
        \begin{tabular}{l|c|cc|ccccc|c|cccc}
            \multirow{2}{*}{{\textbf{Task}}} & \textbf{Num. of} & \multicolumn{2}{c|}{\textbf{Num. of Obj.}} & \multicolumn{5}{c|}{\textbf{Reference Motion Dataset}} & \multirow{2}{*}{\textbf{Epis. Len. (s)}} & \multicolumn{4}{c}{\textbf{Early Termination Condition}} \\
             & \textbf{Task Tokens} & \footnotesize{\textbf{Train}} & 
             \footnotesize{\textbf{Test}} & \footnotesize{\textbf{Loco}} & \footnotesize{\textbf{Stair}} & \footnotesize{\textbf{Climb}} & \footnotesize{\textbf{Carry}} & \footnotesize{\textbf{Sit}} & {} & \footnotesize{\textbf{Char. Fall}} & \footnotesize{\textbf{Obj. Fall}} & \footnotesize{\textbf{Path Dist.}} & \footnotesize{\textbf{IET}}  \\ \hline\hline 
            Follow & 1 & / & / & \checkmark & & & & & 10 & \checkmark & & \checkmark & \\
            Sit & 1 & 49 & 26 &\checkmark & & & & \checkmark & 10 & \checkmark & & & \checkmark \\
            Climb & 1 & 38 & 26 & \checkmark & & \checkmark & & & 10 & \checkmark & & & \checkmark \\
            Carry & 1  & 9 & 9 & \checkmark & & & \checkmark & & 20 & \checkmark & & & \\
            \hline
            Follow + Carry & 3 & / + 5 & / + 9 & \checkmark & & & \checkmark & & 10 & \checkmark & \checkmark & \checkmark & \\
            Sit + Carry & 3 & 49 + 5 & 26 + 9 & \checkmark & & & \checkmark & \checkmark & 10 & \checkmark & \checkmark & & \checkmark \\
            Climb + Carry & 3 & 38 + 5 & 26 + 9 & \checkmark & & \checkmark & \checkmark & & 10 & \checkmark & \checkmark & & \checkmark \\
            \hline
            Obj. Shap. Var. (Chair) & 1 & 63 & 27 & \checkmark & & & \checkmark & & 20 & \checkmark & \checkmark & & \\
            Obj. Shap. Var. (Table) & 1 & 21 & 9 & \checkmark & & & \checkmark & & 20 & \checkmark & \checkmark & & \\
            Terr. Shap. Var. (Follow) & 2 & / & / & \checkmark & \checkmark & & & & 10 & \checkmark & & \checkmark & \\
            Terr. Shap. Var. (Carry) & 2 & 9 & 9 & \checkmark & \checkmark & & \checkmark & & 20 & \checkmark & & & \\
            \hline
            Long-horizon Task & 5 & / & / & \checkmark & & \checkmark & \checkmark & \checkmark & 40 & \checkmark & & & \checkmark
        \end{tabular}
    }
    \caption{The overview of all $12$ tasks implemented in this paper. Key settings for each task are summarized, including the number of task tokens, the construction of reference motion and object datasets, the episode length, and early termination conditions. The available termination conditions contain character fall, object fall, path distance, and interaction early termination (IET). A slash (/) indicates that the specific configuration is not applicable.}
    \label{tab:supp_task_overview}
\end{table*}

\subsection{Preliminaries}
\label{sec:supp_task_pre}

\paragraph{Reference Motion Dataset.} To encourage the character to perform tasks in a realistic and life-like manner, we manually construct a comprehensive reference motion dataset encompassing a wide variety of behavior categories associated with the four foundational HSI tasks. The dataset is divided into five distinct subsets: 

\begin{itemize}
    \item \textbf{Loco:} This subset includes $12$ motion sequences from the AMASS~\cite{mahmood2019amass} dataset, covering basic locomotion behaviors such as standing, walking, and turning around on flat ground. Since every task involves a walking stage, this subset is used for all tasks.

    \item \textbf{Stair:} This subset is used for tasks on stairs terrain and consists of $20$ motion sequences for ascending and descending stairs.

    \item \textbf{Climb}: To support the training of the climbing task, we collect $11$ motion sequences from the AMASS~\cite{mahmood2019amass} dataset, where characters climb onto a high platform from the ground.

    \item \textbf{Carry:} We collect the carrying motions from hybrid sources, with $17$ sequences from the OMOMO~\cite{li2023object} dataset and $4$ sequences from the AMASS~\cite{mahmood2019amass} dataset.
 
    \item \textbf{Sit:} This subset consists of $20$ sitting motions collected from the SAMP~\cite{hassan2021stochastic} dataset.
    
\end{itemize}
The usage of these five motion datasets in each task's training process is summarized in Tab.~\ref{tab:supp_task_overview}. 
For skill composition tasks, such as sitting down while carrying an object, no post-processing is applied to merge the two corresponding subsets (\ie, Carry and Sit) to obtain composite kinematic reference motions. Therefore, the reference motion dataset does not include any motions of sitting down while carrying a box. The policy learns these composite skills primarily through the guidance of task rewards. As the policy learns composite tasks, the style reward decreases while the task reward increases, leading to an overall increase in total reward and improved task completion.

\paragraph{Object Dataset.} To make learned interaction skills effectively generalize to diverse unseen objects, we construct an object training dataset and a corresponding testing dataset to evaluate the generalization capabilities of these skills. The high-quality 3D object models are collected from the 3D-Front~\cite{fu20213d} object dataset, while the 3D models of boxes are procedurally generated using Trimesh. The number of objects used for training and testing is described in Table~\ref{tab:supp_task_overview}. We carefully ensure all test are conducted on the unseen objects.

\paragraph{Early Termination Condition.} Early termination is an effective technique for improving the reinforcement learning (RL) training process by preventing negative samples from adversely affecting the policy gradient~\cite{peng2018deepmimic}. A fundamental and widely applicable termination condition is humanoid fall detection, which is utilized for all tasks in our implementation. To further facilitate the learning of dynamic object-carrying skills, we introduce a similar condition called object fall detection. If the object's height drops below a specified threshold, the trial will be terminated. For the path following task, we adopt the path distance detection condition proposed in Trace\&Pace~\cite{rempe2023trace}. If the 2D distance between the root of the simulated character and the target point on the trajectory at the current moment exceeds a specified threshold, the trial will be terminated. For tasks such as sitting and climbing, where the physical character enters a static interaction state with the object upon task completion, we introduce the Interaction Early Termination (IET) condition proposed by InterScene~\cite{pan2024synthesizing}. This effectively enhances smoothness and increases the success rate performance of these particular tasks.

\subsection{Foundational HSI Tasks} \label{sec:supp_task_foundational}

\subsubsection{Path Following}

\noindent \textbf{Definition.} This task requires the simulated character to move along a target 2D trajectory. We follow the prior work~\cite{rempe2023trace} to procedurally generate the trajectory dataset. A whole trajectory is formulated as $\tau=\{ x_{0.1}^{\tau}, x_{0.2}^{\tau}, ..., x_{T-0.1}^{\tau}, x_{T}^{\tau} \}$, where $x_{0.1}^{\tau}$ denotes a 2D waypoint of the trajectory $\tau$ at the simulation time $0.1s$, and $T$ is the episode length. According to Tab.~\ref{tab:supp_task_overview}, the path following task's episode length $T$ is $10s$. The character needs to follow this trajectory $\tau$ accurately.

\noindent \textbf{Task Observation.} At each simulation moment $t$ second, we query $10$ future waypoints $\{ x_{t}^{\tau}, x_{t+0.1}^{\tau}, ..., x_{t+0.8}^{\tau}, x_{t+0.9}^{\tau} \}$ in the future $1.0s$ from the whole trajectory $\tau$ by linear interpolation. The sampling time interval is $0.1s$. We use the 2D coordinates of the sampled waypoints as the task observation $g_t^{f} \in \mathbb{R}^{2 \times 10}$.

\noindent \textbf{Task Reward.} The task reward $r_t^{f}$ calculates the distance between the current character 2D root position $x_t^{root\_2d}$ and the desired target waypoint $x^{\tau}_t$:
\begin{equation}
    r^f_t = \text{exp} \big ( - 2.0 \left \| x^{root\_2d}_t - x^{\tau}_t \right \|^2  \big ).
    \label{eq:traj_reward}
\end{equation}

\subsubsection{Sitting}

\noindent \textbf{Definition.} The task objective is for the character to move its root joint to a target 3D sitting position located on the object surface. The target position is placed $10$ cm above the center of the top surface of the chair seat.

\noindent \textbf{Task Observation.} The sitting task observation $g_t^{s} \in \mathbb{R}^{38}$ includes the 3D target sitting position $\in \mathbb{R}^3$ and the 3D information of the interacting object, \ie, the root position $\in \mathbb{R}^3$, the root rotation $\in \mathbb{R}^6$, and the 2D front-facing direction $\in \mathbb{R}^2$, as well as the positions of eight corner points on the object's bounding box $\in \mathbb{R}^{3\times8}$. 

\noindent \textbf{Task Reward.} The sitting policy is trained by minimizing the distance between the character's 3D root position $x_t^{root}$ and the target 3D sitting position $x_t^{tar}$.  The task reward $r_t^{s}$ is defined as:
\begin{equation}
    r^s_t=\left\{
    \begin{aligned}
    0.7 \ r_t^{near} &+ 0.3 \ r_t^{far}, \left \| x_t^{obj\_2d} - x^{root\_2d}_t \right \| > 0.5 \\
    0.7 \ r_t^{near} &+ 0.3, \text{otherwise}
    \end{aligned}
    \right.
\label{eq:sit_total_reward}
\end{equation}
\begin{equation}
    \begin{aligned}
    r_t^{far} &= \text{exp} \big ( -2.0 \left \| 1.5 - d_t^* \cdot \dot{x}^{root\_2d}_t \right \|^2 \big )
    \end{aligned}
\label{eq:far_reward}
\end{equation}
\begin{equation}
r_t^{near} = \text{exp}\big(-10.0 \left \| x_t^{tar} - x_t^{root} \right \|^2 \big),
\label{eq:near_reward}
\end{equation}
where $x^{root}_t$ is the 3D coordinates of the character's root,  $\dot{x}^{root\_2d}_t$ is the 2D linear velocity of the character's root, $x^{obj\_2d}_t$ is the 2D position of the object root, $d^*_t$ is a horizontal unit vector pointing from $x^{root\_2d}_t$ to $x^{obj\_2d}_t$, $a \cdot b$ represents vector dot product.

\subsubsection{Climbing}

\noindent \textbf{Definition.} In this work, we introduce a new contact-based interaction task similar to the sitting task. The goal is for the character to stand on a given object, placing its root joint at a target 3D climbing position. We place the target position $94$ cm above the center of the top surface of the object. 

\noindent \textbf{Task Observation.} The task observation $g_t^{m} \in \mathbb{R}^{27}$ includes the target root position $\in \mathbb{R}^3$ and the 3D coordinates of eight corner points on the object's bounding box $\in \mathbb{R}^{3\times8}$.

\noindent \textbf{Task Reward.}  This task is also optimized through minimizing the 3D distance between the character's root $x_t^{root}$ and its target location $x_t^{tar}$. We formulate the task reward $r_t^{m}$, as follows:
\begin{equation}
    r^m_t=\left\{
    \begin{aligned}
    0.5 \ r_t^{near} &+ 0.2 \ r_t^{far}, \left \| x_t^{obj\_2d} - x^{root\_2d}_t \right \| > 0.7 \\
    0.5 \ r_t^{near} &+ 0.2 + 0.3 \ r_t^{foot}, \text{otherwise}
    \end{aligned}
    \right.
\label{eq:climb_total_reward}
\end{equation}
\begin{equation}
    \begin{aligned}
    r_t^{far} &= \text{exp} \big ( -2.0 \left \| 1.5 - d_t^* \cdot \dot{x}^{root\_2d}_t \right \|^2 \big )
    \end{aligned}
\label{eq:far_reward}
\end{equation}
\begin{equation}
r_t^{near} = \text{exp}\big(-10.0 \left \| x_t^{tar} - x_t^{root} \right \|^2 \big)
\label{eq:near_reward}
\end{equation}
\begin{equation}
r_t^{foot} = \text{exp}\big(-50.0 \left \| (x_t^{tar\_h} - 0.94) - x_t^{foot\_h} \right \|^2 \big),
\label{eq:foot_reward}
\end{equation}
where $x_t^{tar\_h}$ denotes the height component of the 3D target root position, $x_t^{tar}$, $(x_t^{tar\_h} - 0.94)$ represent the height of the top surface of the target object in the world coordinate, and $x_t^{foot\_h}$ denotes the mean height of the two foot rigid bodies. The reward function $r_t^{foot}$ is introduced to encourage the character to lift its feet, which is applied when the character is close enough to the target object. We find it is crucial for the successful training of the climbing task.

\subsubsection{Carrying}

\noindent \textbf{Definition.} The character is directed to move a box from a randomly initial 3D location $x_t^{box\_init}$ to a target 3D location $x_t^{box\_tar}$. We use two thin platforms to support the box since the its initial and target heights are randomly generated.

\noindent \textbf{Task Observation.} The task observation $g_t^{c} \in \mathbb{R}^{42}$ comprises the following properties of the target box:
\begin{itemize}
    \item Target location of the box $\in \mathbb{R}^3$
    \item Root position $\in \mathbb{R}^3$
    \item Root rotation $ \in \mathbb{R}^6$
    \item Root linear velocity $\in \mathbb{R}^3$ 
    \item Root angular velocity $\in \mathbb{R}^3$ 
    \item Positions of $8$ corner points on the bounding box $\in \mathbb{R}^{3 \times 8}$ 
\end{itemize}

\noindent \textbf{Task Reward.} We implement the multi-stage task reward function proposed by InterPhys~\cite{hassan2023synthesizing}. The first stage aims to encourage the character to walk to the initial box. The corresponding reward $r_t^{c\_walk}$  is defined as:
\begin{equation}
    r^{c\_{walk}}_t=\left\{
    \begin{aligned}
    &0.2 , \left \| x_t^{obj\_2d} - x^{root\_2d}_t \right \| < 0.5 \\
    &0.2 \ \text{exp} \big ( -5.0 \left \| 1.5 - d_t^* \cdot \dot{x}^{root\_2d}_t \right \|^2 \big ) , \\ & 
 \ \ \ \  \ \ \ \ \ \ \ \ \ \ \  \ \ \ \ \ \ \   \ \ \ \  \ \ \ \ \ \ \   \ \ \ \  \ \ \ \ \ \ \  \text{otherwise}
    \end{aligned}
    \right.
\label{eq:carry_s1}
\end{equation}
where $d^*_t$ is a horizontal unit vector pointing from $x^{root\_2d}_t$ to $x^{obj\_2d}_t$, $a \cdot b$ represents vector dot product. The second stage is to encourage the character to pick up and move the box to its target location. We utilize two reward functions to achieve this stage, \ie, $r^{c\_{carry}}_t$ to calculate the 3D distance between the box current root position $x_t^{obj}$ and its target location $x_t^{tar}$, and $r^{c\_{pick}}_t$ to calculate the 3D distance between the box position $x_t^{obj}$ and the mean 3D position of the character's two hands $x^{hand}$. We define  $r^{c\_{carry}}_t$ as follows:
\begin{equation}
    r^{c\_carry}_t=\left\{
    \begin{aligned}
    0.2 \ r_t^{near} &+ 0.2 \ r_t^{far},  \\ 
    & \left \| x_t^{obj\_2d} - x^{tar\_2d}_t \right \| > 0.5 \\
    0.2 \ r_t^{near} &+ 0.2, \text{otherwise}
    \end{aligned}
    \right.
\label{eq:carry_s2_carry_total_reward}
\end{equation}
\begin{equation}
    \begin{aligned}
    r_t^{far} &= \text{exp} \big ( -5.0 \left \| 1.5 - d_t^\# \cdot \dot{x}^{obj\_2d}_t \right \|^2 \big )
    \end{aligned}
\label{eq:eq:carry_s2_carry_far}
\end{equation}
\begin{equation}
r_t^{near} = \text{exp}\big(-10.0 \left \| x_t^{tar} - x_t^{obj} \right \|^2 \big),
\label{eq:carry_s2_carry_near}
\end{equation}
where $x^{obj}_t$ is the 3D coordinates of the box's root,  $\dot{x}^{obj\_2d}_t$ is the 2D linear velocity of the box's root, $x^{obj\_2d}_t$ is the 2D position of the object root, $x^{tar\_2d}_t$ is the 2D coordinates of the box's target location, $d^\#_t$ is a horizontal unit vector pointing from $x^{obj\_2d}_t$ to $x^{tar\_2d}_t$, $a \cdot b$ represents vector dot product. The task reward $r^{c\_pick}_t$ to incentivize the character pick up the box using its hands, defined as follows:
\begin{equation}
    r^{c\_{pick}}_t=\left\{
    \begin{aligned}
    &0.0 , \left \| x_t^{obj\_2d} - x^{root\_2d}_t \right \| > 0.7 \\
    &0.2 \ \text{exp}\big(-5.0 \left \| x_t^{obj} - x_t^{hand} \right \|^2 \big) , \\ & 
 \ \ \ \  \ \ \ \ \ \ \ \ \ \ \  \ \ \ \ \ \ \   \ \ \ \  \ \ \ \ \ \ \   \ \ \ \  \ \ \ \ \ \ \  \text{otherwise}
    \end{aligned}
    \right.
\label{eq:carry_s2_pick}
\end{equation}
where $x_t^{hand}$ denotes the mean 3D coordinates of the character's two hands. Additionally, we further design a reward function $r_t^{c\_put}$ to incentivize the character to put down the box at its target location accurately, which is formulated as:
\begin{equation}
    r^{c\_{put}}_t=\left\{
    \begin{aligned}
    &0.0 , \left \| x_t^{obj\_2d} - x^{tar\_2d}_t \right \| > 0.1 \\
    &0.2 \ \text{exp}\big(-10.0 \left \| x_t^{obj\_h} - x_t^{tar\_h} \right \|^2 \big) , \\ & 
 \ \ \ \  \ \ \ \ \ \ \ \ \ \ \  \ \ \ \ \ \ \   \ \ \ \  \ \ \ \ \ \ \   \ \ \ \  \ \ \ \ \ \ \  \text{otherwise}
    \end{aligned}
    \right.
\label{eq:carry_s3_put}
\end{equation}
where $x_t^{obj\_h}$ denotes the hight of the current box and $x_t^{tar_h}$ represents the height of the target placing position. Therefore, the total reward function $r_t^c$ for training the carrying skill can be formulated as:
\begin{equation}
    r^{c}_t=r^{c\_walk}_t + r^{c\_carry}_t + r^{c\_pick}_t + r^{c\_put}_t.
\label{eq:carry_total}
\end{equation}

\subsection{Downstream HSI Tasks} \label{sec:supp_task_advanced}

In this section, we provide the details about how we implement the task observations and rewards used for training these more challenging HSI tasks.

\noindent \textbf{Skill Composition.} When learning the composite tasks using our policy adaptation, we reuse and freeze two relevant task tokenizers of foundational skills. Their task observations are illustrated in Sec~\ref{sec:supp_task_foundational}. Therefore, we mainly focus on describing how we construct the task configurations for these composite tasks.

\textit{Follow + Carry}. The task observation $g_t^{f+c}$ contains two parts: (1) the primary following task observation $g_t^{f} \in \mathbb{R}^{2\times10}$ and (2) the revised carrying task observation $g_t^{c\_{revised}} \in \mathbb{R}^{39}$, which excludes the target location of the box $\in \mathbb{R}^{3}$ because the carrying task is no longer the primary task. The final composite task observation is $g_t^{f+c} \in \mathbb{R}^{2\times10+39} $ that considers both the primary task states and the carrying box states. We define the task reward for this composite task  $r_t^{f+c}$ as follows:
\begin{equation}
    r^{f+c}_t=\left\{
    \begin{aligned}
    & 0.0, \left \| x_t^{obj\_2d} - x^{root\_2d}_t \right \| > 0.7 \\
    & 0.5 \ r_t^{f} + 0.5 \ r_t^{c\_pick}, \text{otherwise}
    \end{aligned}
    \right.
 \label{eq:follow+carry_total}
\end{equation}
where the $r_t^{f}$ is the same as Equ.~\ref{eq:traj_reward} and the $r_t^{c\_pick}$ is equal to Equ.~\ref{eq:carry_s2_pick}.

\textit{Sit + Carry}. The task observation $g_t^{s+c}$ also includes two parts: (1) the primary sitting task observation $g_t^{s} \in \mathbb{R}^{38}$ and (2) the revised carrying task observation $g_t^{c\_{revised}} \in \mathbb{R}^{39}$, which have been introduced before. The final composite task observation is $g_t^{s+c} \in \mathbb{R}^{38+39} $. We define the task reward for this composite task  $r_t^{s+c}$ as follows:
\begin{equation}
    r^{s+c}_t=\left\{
    \begin{aligned}
    & 0.0, \left \| x_t^{obj\_2d} - x^{root\_2d}_t \right \| > 0.7 \\
    & 0.7 \ r_t^{s} + 0.3 \ r_t^{c\_pick}, \text{otherwise}
    \end{aligned}
    \right.
\label{eq:sit+carry_total}
\end{equation}
where the $r_t^{s}$ is the same as Equ.~\ref{eq:sit_total_reward} and the $r_t^{c\_pick}$ is equal to Equ.~\ref{eq:carry_s2_pick}.

\textit{Climb + Carry}. The task observation $g_t^{m+c}$ also includes two parts: (1) the primary climbing task observation $g_t^{m} \in \mathbb{R}^{27}$ and (2) the revised carrying task observation $g_t^{c\_{revised}} \in \mathbb{R}^{39}$, which have been introduced before. The final composite task observation is $g_t^{m+c} \in \mathbb{R}^{27+39} $. We define the task reward for this composite task  $r_t^{m+c}$ as follows:
\begin{equation}
    r^{m+c}_t=\left\{
    \begin{aligned}
    & 0.0, \left \| x_t^{obj\_2d} - x^{root\_2d}_t \right \| > 0.7 \\
    & 0.7 \ r_t^{m} + 0.3 \ r_t^{c\_pick}, \text{otherwise}
    \end{aligned}
    \right.
\label{eq:sit+carry_total}
\end{equation}
where the $r_t^{m}$ is the same as Equ.~\ref{eq:climb_total_reward} and the $r_t^{c\_pick}$ is equal to Equ.~\ref{eq:carry_s2_pick}.

\noindent \textbf{Object/Terrain Shape Variation.} For object shape variation, we directly fine-tune the pre-trained box-carrying task tokenizer $\mathbb{T}^{c}$. That is, the task observation is still $g_t^{c} \in \mathbb{R}^{42}$. And we reuse the box-carrying reward function Equ.~\ref{eq:carry_total}. For terrain shape variation, we introduce an additional task tokenizer for perceiving the surrounding height map, which use $1024$ sensor points to represent the height values in a $2 \times 2 \ m^2$ square area centered at the humanoid root position. Thus, the new height map observation is $g_t^{new} \in \mathbb{R}^{1024}$. We also reuse the task observation and reward function of the box-carrying task for terrain shape variation.


\noindent \textbf{Long-horizon Task Completion.} As illustrated in Fig.~\ref{fig:result} (g), we sequence the four learned foundational skills to perform a long-horizon task in a complex environment. We reuse all observations of foundational skills and introduce a new height map observation $g_t^{new} \in \mathbb{R}^{625}$, which utilizes $625$ sensor points to observe the heights in a $1 \times 1 \ m^2$ square area. We design a step-by-step task reward mechanism. For each step in the task sequence, we reuse the task reward from the corresponding task ($(r_t^f, r_t^s, r_t^m$, or $r_t^c)$). Once a step is completed, the reward value for that sub-task is set to its maximal value, indicating the task have been accomplished. Then, the task reward for the next step in the sequence will be activated for reward calculation. Please refer to our publicly released code for more details.

\section{Implementation Details of CML} \label{sec: detailed_cml}


In Sec.~\ref{exp:comp}, we compare our transformer-based policy adaptation with CML~\cite{xu2023composite} and CML (dual) on the skill composition tasks. The network structure of CML (dual) is an improved version based on the original CML framework.

CML~\cite{xu2023composite} employs a hierarchical framework consisting of a pre-trained, fixed meta policy $\pi^{meta}$ as the low-level controller and a newly introduced, trainable policy $\pi^{new}$ as the high-level controller. The high-level policy $\pi^{new}$ observes the humanoid proprioception $s_t$ and the new task observation $g_t^{new}$. The low-level policy $\pi^{meta}$ observes the humanoid proprioception $s_t$ and the base task observation $g_t^{base}$. Take the \textit{Climb + Carry} task as an example. We use a specialist policy trained on the climbing task as the $\pi^{meta}(a_t^{meta}|s_t,g_t^m)$, which possesses a joint character-goal state space. Then, we introduce a new policy $\pi^{new}(a_t^{new}, w_t^{new}|s_t,g_t^{m+c})$, which generates a new action $a_t^{new}$ and a group of joint-wise weights $w_t^{new} \in \mathbb{R}^{32}$, each value is $\in [0, 1]$. The high-level policy $\pi^{new}$ is trained to cooperate with the low-level policy $\pi^{meta}$ to quickly learn the composite tasks. The composition process is conducted in the action space as follows:
\begin{equation}
    a_t = a_t^{new} + w_t^{new} a_t^{meta},
\label{eq:cml}
\end{equation}
which is called post-composition in the main paper.

However, the original CML framework supports only a single meta policy. To ensure a fair comparison, we develop CML (dual), an improved version that can simultaneously utilize two meta policies $\pi^{meta}_1$ and $\pi^{meta}_2$. To handle the two sets of actions $a_t^{meta_1}$ and $a_t^{meta_2}$, generated by the two meta policies, the high-level policy $\pi^{new}$ outputs an additional set of weights. In this way, we obtain $w_t^{new_1}$ and $w_t^{new_2}$. This results in the following post-composition process:
\begin{equation}
    a_t = a_t^{new} + w_t^{new_1} a_t^{meta_1} + w_t^{new_2} a_t^{meta_2},
\label{eq:cml_dual}
\end{equation}
where $w_t^{new_1}$ and $w_t^{new_2}$ are joint-wise weights applied to the two sets of meta actions, $a_t^{meta_1}$ and $a_t^{meta_2}$, respectively. All weights are processed using sigmoid activations, transforming their values to $[0, 1]$.

\begin{figure}[t]
  \centering
   \includegraphics[width=0.9\linewidth]{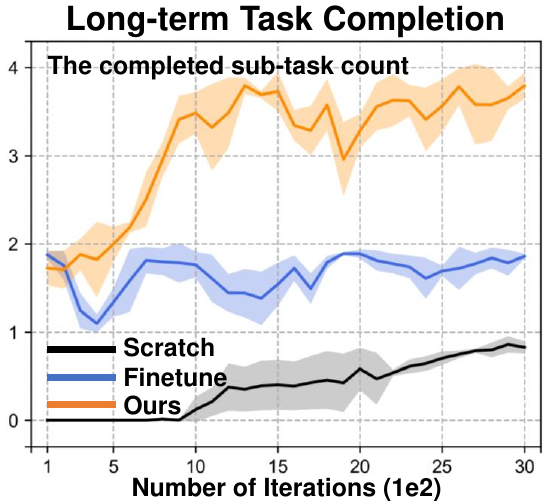}  
   \caption{Learning curves comparing the efficiency on long-horizon task completion using \ours, Scratch~\cite{peng2021amp}, and iterative fine-tuning of multiple pre-trained specialist policies, namely Finetune.}
   \label{fig:supp_long_efficiency}
\end{figure}

\section{Quantitative Evaluation on Long-horizon Task Completion} \label{sec:supp_quan_long}


\noindent \textbf{Experimental Setup.} We first describe the construction of the long-horizon task shown in Fig.~\ref{fig:result} (g). The long task comprises four sequential sub-tasks: follow a target trajectory $\rightarrow$ carry a box to its target location $\rightarrow$ climb onto the box $\rightarrow$ sit on a chair located on the high platform. Each sub-task should have a sub-goal. The finite state machine monitors the task executing process using the spatial relationship between the reference point (the humanoid root joint or the box centroid) and the sub-goal. Each sub-goal is procedurally generated—follow: the trajectory is planned by A*; carry: the target box position is placed close to the platform using a rule-based method; climb and sit: the character’s target root position is pre-defined on the object geometry. We use the completed sub-task count as the evaluation metric. The maximal value is $4$ in our case. We collect $512$ trials to statistic the metrics.

\noindent \textbf{Baselines.} We compare our approach with two baseline methods: (1) Scratch~\cite{peng2021amp}: training a policy to learn the whole long-term task from scratch; (2) Finetune~\cite{clegg2018learning,leeadversarial}: iterative fine-tuning multiple specialist policies in the environment to improve skill transitions and collision avoidance. We use our transformer policy as the policy architecture when conducting the experiment Scratch. Both Scratch and \ours can observe height map. The difference between these two approaches is that our approach utilized pre-trained parameters. Due to the limited flexibility of MLP-based policies used by the experiment Finetune, we cannot make them environment-aware. The training adopts $1, 024$ parallel simulation environments and $3k$ PPO iterations.

\noindent \textbf{Quantitative Results.} Our method achieves the highest value of the completed sub-task count of $3.79\pm0.14$, significantly outperforming Scratch $0.82\pm0.06$ and Finetune $1.86\pm0.02$. We also illustrate the convergence curves in Fig.~\ref{fig:supp_long_efficiency}, which shows that \ours still maintains its efficiency advantage in the long-horizon task completion.

\begin{figure*}[t]
  \centering
   \includegraphics[width=1.0\linewidth]{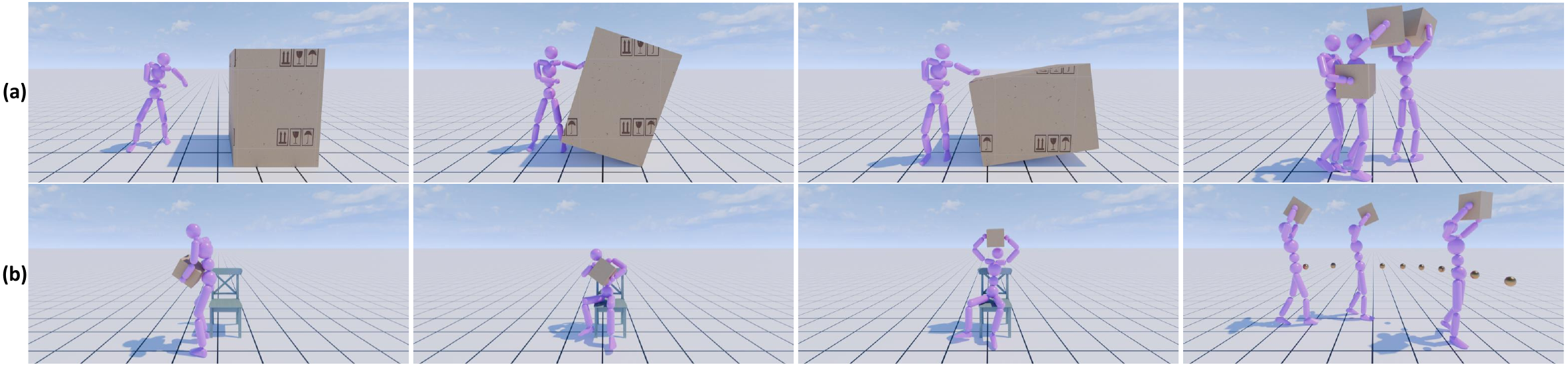}  
   \caption{Qualitative results of new skills learned by our policy adaptation. (a) We first learn two out-of-domain interaction skills, \ie, pushing down a large object and walking to a target location while lifting up a box using two hands. (b) We then combine the new lifting skill with previously learned sitting and path-following skills. These results demonstrate the good extensibility of our transformer policy.}
   \label{fig:supp_new_skills}
\end{figure*}

\section{Extensibility} \label{sec:supp_extensibiliy}

In the main paper, we mainly focus on adapting skills learned in the first stage (\ie, foundational skill learning) to address more challenging HSI tasks through policy adaptation. In this section, we want to evaluate the extensibility of our approach to more HSI skills. We attempt to answer two questions: (1) Can we insert out-of-domain skills into the pre-trained transformer policy? (2) Can we further combine the newly-added skills with previously learned foundational skills to create more compositional cases? The flexibility of the transformer policy allows us to explore these problems.

\noindent \textbf{Q1: Out-of-domain Skill Insertion.} We consider two more types of manipulation skills, including pushing down a large object and walking to a target location while lifting up a box using two hands. To prepare the training, we collect the reference motions from the AMASS dataset~\cite{mahmood2019amass}, design the task observations, rewards, and other environmental configurations. During training, we introduce a randomly initialized task tokenizer $\mathbb{T}^{new}$ and zero-initialized adapter layers to the action head $\mathbb{H}$. The rest network parameters are frozen. For the pushing task, we declare it to be successful if the object falls. For the lifting task, we determine a testing trial to be successful if the pelvis is within $20$ cm (XY-planar distance) of the target location while maintaining the box in a lifted position. As shown in Fig.~\ref{fig:supp_new_skills} (a), our approach successfully synthesizes these out-of-domain manipulation skills. Specifically, the pushing task attains a success rate of 100\% and the lifting task receives $80.6\%\pm2.6$.

\noindent \textbf{Q2: More Compositional Cases.} Moreover, we combine the newly learned box-lifting skill with previous sitting and following skills. The training method is the same as skill composition. Through policy adaptation, we create more compositional cases shown in Fig.~\ref{fig:supp_new_skills} (b). The success rates are $72.1\%\pm2.8$ and $91.1\%\pm0.8$, respectively.


\end{document}